\documentclass[lettersize,journal]{IEEEtran}
\usepackage{amsmath,amsfonts}
\usepackage{algorithmic}
\usepackage{algorithm}
\usepackage{array}
\usepackage[caption=false,font=normalsize,labelfont=sf,textfont=sf]{subfig}
\usepackage{textcomp}
\usepackage{stfloats}
\usepackage{url}
\usepackage{pifont}
\usepackage{verbatim}
\usepackage{graphicx}
\usepackage{cite}
\usepackage{enumitem}
\usepackage[table,xcdraw]{xcolor}
\usepackage{hyperref}
\hypersetup{
    colorlinks=true,    
    linkcolor=blue,     
    citecolor=blue,     
    urlcolor=blue       
}
\usepackage{cleveref} 
\usepackage{multirow}
\usepackage{tabularx}
\usepackage{booktabs}
\usepackage{makecell}
\usepackage{amssymb}
\usepackage{orcidlink}

\Crefname{figure}{fig.}{figs.}
\Crefname{figure}{Fig.}{Figs.}

\Crefname{equation}{Eq.}{Eqs.}

\newcommand{\thickhline}{\noalign{\hrule height 0.8pt}}

\definecolor{myred}{HTML}{FF1919}
\definecolor{mygreen}{HTML}{A9D18E}
\definecolor{myorange}{HTML}{FF9900}

\begin{document}

\title{ Single-Point Supervised High-Resolution Dynamic Network for Infrared Small Target Detection}

\author{Jing Wu\hspace{-1.5mm}$^{~\orcidlink{0000-0002-0181-4095}}$, Rixiang Ni\hspace{-1.5mm}$^{~\orcidlink{0009-0007-4614-9977}}$, Feng Huang\hspace{-1.5mm}$^{~\orcidlink{0000-0003-4652-4312}}$, Zhaobing Qiu\hspace{-1.5mm}$^{~\orcidlink{0000-0003-3834-3802}}$, Liqiong Chen\hspace{-1.5mm}$^{~\orcidlink{0000-0001-5058-0523}}$, Changhai Luo\hspace{-1.5mm}$^{~\orcidlink{0009-0004-4489-3354}}$, 

Yunxiang Li\hspace{-1.5mm}$^{~\orcidlink{0009-0002-9422-1666}}$, and Youli Li\hspace{-1.5mm}$^{~\orcidlink{0009-0009-9080-577X}}$

\thanks{This work was supported in part by the Nature Science Foundation
of Fujian Province under Grant 2022J05113, and in part by the Educational
Research Program for Young and Middle-aged Teachers of Fujian Province
under Grant JAT210035. (\textit{Jing Wu and Rixiang Ni contributed
 equally to this work.}) (\textit{Corresponding author: Feng Huang; Zhaobing Qiu.})
 
Jing Wu, Rixiang Ni, Feng Huang, Zhaobing Qiu, Liqiong Chen, Changhai Luo, Yunxiang Li, Youli Li are with the School of Mechanical Engineering and Automation, Fuzhou University, Fuzhou 305108, China 

}

}

\maketitle
\markboth{ IEEE TRANSACTIONS ON GEOSCIENCE AND REMOTE SENSING}%
{Shell \MakeLowercase{\textit{et al.}}: A Sample Article Using IEEEtran.cls for IEEE Journals}

\begin{abstract}
Infrared small target detection (IRSTD) tasks are extremely challenging for two main reasons: 1) it is difficult to obtain accurate labelling information that is critical to existing methods, and 2) infrared (IR) small target information is easily lost in deep networks. To address these issues, we propose a single-point supervised high-resolution dynamic network (SSHD-Net). In contrast to existing methods, we achieve state-of-the-art (SOTA) detection performance using only single-point supervision. Specifically, we first design a high-resolution cross-feature extraction module (HCEM), that achieves bi-directional feature interaction through stepped feature cascade channels (SFCC). It balances network depth and feature resolution to maintain deep IR small-target information. Secondly, the effective integration of global and local features is achieved through the dynamic coordinate fusion module (DCFM), which enhances the anti-interference ability in complex backgrounds. In addition, we introduce the high-resolution multilevel residual module (HMRM) to enhance the semantic information extraction capability. 
Finally, we design the adaptive target localization detection head (ATLDH) to improve detection accuracy. 
Experiments on the publicly available datasets NUDT-SIRST and IRSTD-1k demonstrate the effectiveness of our method. Compared to other SOTA methods, our method can achieve better detection performance with only a single point of supervision.
\end{abstract}


\section{Introduction}
\IEEEPARstart{I}{nfrared} small target detection (IRSTD) technology is widely used in several fields, including infrared early warning \cite{Missile_Defense_RLPGB_2}\cite{sun2020infrared_EFL_3} and maritime search \cite{ying2022mocopnet_DNA_2}\cite{teutsch2010classification_DNA_1}, because of its high penetration and stealth. Limited by the properties of IR imaging, resolution, and the need for early warning by detection systems \cite{wang2023rlpgb}, IR small-target usually lack shape and texture features. The characteristics of IR small-target include: 1) small: due to the long imaging distance, the target occupies a small percentage of the IR image, usually consisting of one to less than eighty pixels \cite{zhang2003algorithms_RLPGB_10}; 2) shapeless\cite{li2022DNA}: IR small-target do not have fixed shape influenced by thermal-effect of infrared radiation\cite{zhang2014thermaleffectofinfrared}; 3) low signal-to-clutter ratio (SCR)\cite{yang2024eflnet}: IR small-target are frequently disturbed by complex background noise, making them susceptible to being overwhelmed or obscured; 4) target sparsity\cite{yang2024eflnet}: there are usually a few targets in IRSTD, resulting in a significant imbalance between the target areas and the background. These factors pose significant challenges for IRSTD.

In order to achieve the detection of IR small-target, numerous model-based (M-based) methods have been proposed by researchers. These methods can be divided into three categories depending on the principle of realization: filter-based, \cite{deshpande1999max_efl4}\cite{wang2020wavelet_qiu86}, local contrast-based \cite{hussain1995infrared_qiu3}\cite{wei2016MPCM_qiu110}\cite{qin2019FKRW_qiu120}\cite{qiu2020adaptive_qiu_Self1}, and low-rank-based \cite{gao2013IPI_efl8}\cite{dai2017reweighted_efl23}\cite{zhang2019PSTNN_efl24}. Although M-based methods can improve the detection accuracy, they rely heavily on artificial prior features and struggle with complex detection scenarios.

As deep learning technology improves continuously, deep-learning-based (DL-based) methods can effectively achieve feature learning of IR small-target with the help of powerful data analysis and modeling capabilities. Currently, DL-based methods primarily include segmentation-based methods, \cite{dai2021alcnet}\cite{li2022DNA}\cite{liu2023infrared_kaiti39}\cite{chen2022irstformer_efl29} such as missed detection versus false alarm conditional adversarial network (MDvsFA-cGan) \cite{wang2019MDvsFA_efl10} and asymmetric contextual modulation Network (ACM-Net)\cite{dai2021acm_efl11}, as well as detection-based methods, exemplified by one-stage cascade refinement network (OSCAR) \cite{dai2023oscar_self} and enhancing feature learning network (EFL-Net)\cite{yang2024eflnet}. Although existing methods yield better results, several challenges remain. On the one hand, the edges of the target and the background are usually blurred by thermal-effect of infrared radiation, making it difficult to obtain accurate labelling information. This increases manual costs \cite{ying2023LESPS} and hinders improvements in the network performance. On the other hand, the process of generating deep semantic information is usually accompanied by significant downsampling of the feature map, which leads to the loss of spatially detailed features of IR small-target. In contrast, reducing the network depth prevents us from extracting sufficient target semantic information to overcome complex clutter interference.

To address these issues, we propose a single-point supervised high-resolution dynamic network (SSHD-Net). Notably, we rely solely on single-point labels, marked near the center of the target mask, as the supervisory signals for network training. The SSHD-Net consists of four main components: a high-resolution cross-feature extraction module (HCEM), two dynamic coordinate fusion module (DCFM), a high-resolution multilevel residual module (HMRM), and an adaptive target localization detection head (ATLDH).

Specifically, we design an HCEM with a stepped feature cascade channel (SFCC) for adaptive contextual feature extraction and bi-directional feature interaction. The HCEM achieves a balance between network depth and feature layer resolution in extracting IR small-target semantic features by propagating high- and low-resolution feature layers in parallel. Different resolution feature layers are connected across resolutions at specific nodes through the SFCC, which prompts global and detail information to guide each other. In addition, we design a dynamic coordinate fusion module (DCFM) to extract effective features through an omni-dimensional dynamic convolutional block (ODBlock) and adjust the spatial distribution of the feature maps through a coordinate attention (CA) module to enhance the target localization capability. Next, we design a high-resolution multilevel residual module (HMRM) with a nested residual network structure to extract the semantic information of deep IR small-target effectively. Finally, to reduce manual labelling cost and overcome the labelling uncertainty, we designed an innovative adaptive target localization detection head (ATLDH) using point-to-point target regression. This method achieves target detection by predicting the probabilistic heatmap of a target with a Gaussian distribution and applying adaptive non-maximal suppression (ANMS).

The test results on two public datasets, NUDT-SIRST \cite{li2022DNA} and IRSTD-1k \cite{zhang2022isnet_msaff16}, show that our method outperforms the SOTA methods, which demonstrates the efficiency and innovation of our method. In addition, to promote the development of IRSTD, we provide a centroid-labelled version of the existing public dataset of IR small-target to support more efficient and reliable IR small-target detection studies. The contributions of this study can be summarized as follows:

\setlist[itemize]{leftmargin=2em}
\begin{itemize}
  \item{} We propose a novel SSHD network that effectively improves the detection performance of IR small-target using only single-point supervised labelling.
  \item{} A high-resolution cross-feature extraction module (HCEM) that effectively achieves adaptive contextual feature extraction and bi-directional feature enhancement through stepped feature cascade channels (SFCC). In addition, to improve the efficiency of multi-scale feature fusion and the localization ability of IR small-target, we design a dynamic coordinate fusion module (DCFM).
  \item{}Finally, we design an adaptive target localization detection head with point-to-point target regression, which effectively overcomes the uncertainty of manual annotation and enables the model to focus on the target core region. We also provide a single-point labelled annotated version of the existing public dataset of IR small-target, pushing the field toward more efficient and accurate development.

\end{itemize}

The remainder of this paper is organized as follows: In \Cref{sec:Related Work}, we briefly review related work. In \Cref{sec:Methodology}, we describe the architecture of the proposed SSHD-Net in detail. \Cref{sec:Experiment} presents the experimental results and analyses. Finally, \Cref{sec:Conclusion} presents the conclusions.

\section{Related Work}
\label{sec:Related Work}

\begin{figure*}[!t] 
    \centering
    \includegraphics[width=\textwidth]{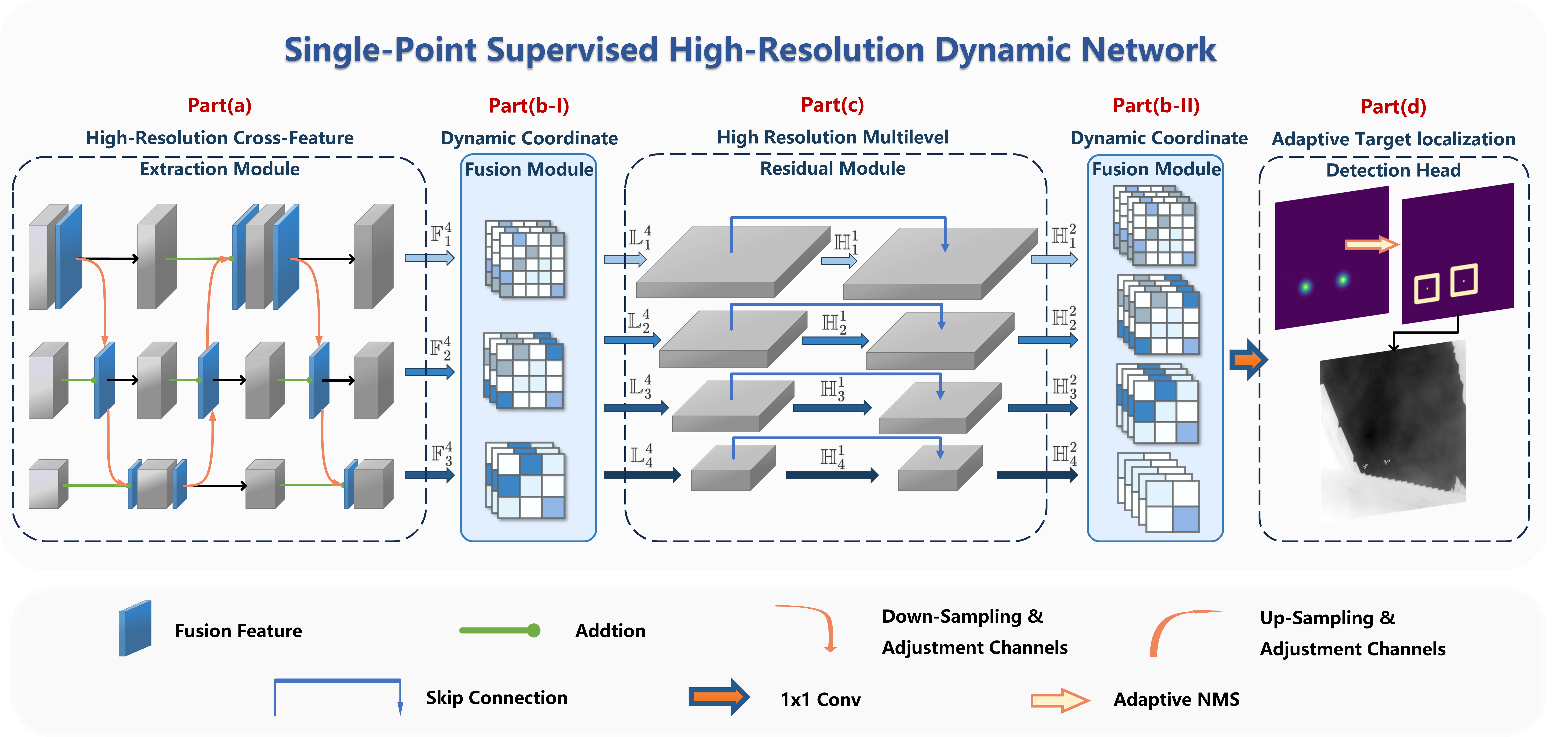}
    \caption{An illustration of the proposed single-point supervised high-resolution dynamic network (SSHD-Net). (a) High-resolution cross-feature extraction module (HCEM). IR images are processed through the Stem layer, downsampled at different ratios, then fed into the HCEM module for feature extraction. (b) Dynamic coordinate attention module (DCFM). The DCFM achieves dynamic enhancement and feature fusion of multi-resolution features in Part b-I and Part b-II. (c) High-resolution multilevel residual module (HMRM). The HMRM extracts deep semantic information of the targets. (d) Adaptive Target localization Detection Head (ATLDH). The features of part b-II output a target localization heatmap by $1\times1$ convolution, which is subsequently processed by ATLDH to obtain the target centroid and confidence.}
    \label{fig:Overall_Architecture}
\end{figure*}

\subsection{Model-Based Methods}
The M-based method focuses on analyzing the prior characteristics of small targets and modeling them according to the differences between the characteristics of the target and background. These methods include filter-, local contrast-, and low-rank-based methods. Filter-based methods achieve IRSTD by designing spatial- or frequency-domain filters to suppress background noise, such as the max-median \cite{deshpande1999max_efl4}, top-hat\cite{bai2010tophat_rlpgb59}, non-subsampled shearlet transform (NSST), and wavelet-based contourlet transform (WBCT)\cite{wang2020wavelet_qiu86}. These methods are simple in principle; however, they rely excessively on prior assumptions. They are susceptible to missed detections in complex backgrounds and noise interference. Local contrast-based methods characterize the differences in the properties of a potential target region from the surrounding background using specific local feature extraction models and representations. These methods include the improved local contrast measure (ILCM) \cite{hussain1995infrared_qiu3}, multi-scale patch- based contrast measure (MPCM)\cite{wei2016MPCM_qiu110}, pixel-level local contrast measure (PLLCM)\cite{qiu2022PLLCM}, and robust unsupervised multi-Feature representation (RUMFR)\cite{chen2024rumfr}. These methods have good real-time performance and interpretability, but the feature extraction and representation of the model is insufficient, which leads to poor model adaptation. Low-rank-based methods use sparse and low-rank matrices to represent the target and background, respectively. It uses component analysis to decompose and the reconstruct matrices. Representative models include the infrared patch-image model (IPI)\cite{gao2013IPI_efl8}, reweighted infrared patch tensor (RIPT)\cite{dai2017reweighted_efl23}, and partial sum of the tensor nuclear norm (PSTNN)\cite{zhang2019PSTNN_efl24}. These methods can improve the accuracy of IRSTD, but may also lead to false alarms under sparse interference.

\subsection{Deep-Learning-Based Methods}
Recently, owing to the continuous progress in deep learning technology, DL-based methods have made impressive progress in IRSTD. These DL-based methods can leverage their powerful data analysis and modelling capabilities to autonomously extract the effective features of IR small-target, significantly improving the precision of detection. These methods can be divided into two main categories: segmentation- and detection-based.

\begin{enumerate}[wide]
\item{\textit{Segmentation-Based Methods}:}
The Segmentation-based methods achieve the localization of IR small-target by classifying every pixel and generating a segmentation mask that contains the target location and shape information. Wang et al. use generative adversarial networks (GAN)\cite{wang2019MDvsFA_efl10}\cite{mirza2014cGan} to decompose a target segmentation task into a question of the balance between missed detections and false alarms. A asymmetric contextual modulation (ACM)\cite{dai2021acm_efl11} developed by Dai et al. combines both top-down and bottom-up point-wise attentional mechanisms to achieve excellent network performance with fewer parameters. Dai et al. also propose attentional local contrast networks (ALC-Net)\cite{dai2021alcnet}, which combine traditional contrast methods with deep learning, demonstrating the potential for the fusion of M-based and DL-based methods. To reduce the information loss of IR small-target in the deep network, Li et al. propose a dense nested attention network (DNA-Net) \cite{li2022DNA}, which improves feature extraction and deep semantic information extraction of targets through dense connectivity and spatial pyramid fusion. Liu et al. \cite{liu2023infrared_kaiti39} introduce transformer structure, which is popular in computer vision and natural language processing, into IRSTD and obtain significant detection results. In addition, Chen et al\cite{chen2022irstformer_efl29} design a hierarchical vision transformer (Irstformer) instead of traditional convolutional kernel to achieve multi-scale feature coding, which effectively solves the problem of long distance dependencies in images.
\item{\textit{Detection-Based Methods}:}
The Detection-based methods achieve IRSTD by regressing the bounding box of the target. Du et al.\cite{du2021spatial_kaiti31} use a mini-IoU strategy on the RPN-ROI detection framework to reduce the misclassification of samples and introduce a new enhancement mechanism to suppress clutter. To improve the efficiency of IRSTD, Ju et al.\cite{ju2021istdet_kaiti32} propose a one-stage detector, ISTDet, which combines an image-filtering module and a detection module to achieve accurate localization. Zhou et al.\cite{zhou2022yolo_kaiti33} integrate the SASE module into a multilevel YOLO network, which significantly improves the robustness of the model for complex backgrounds. Dai et al.\cite{dai2023oscar_self} propose a single-stage cascade refinement network (OSCAR) to improve the regression accuracy of the bounding box by performing target regression from coarse to fine. In addition, Yang et al. \cite{yang2024eflnet} design an enhancing feature learning network (EFL-Net) to address the extreme imbalance between the target and background and the high sensitivity of small infrared targets to bounding-box regression.

Although these methods are significantly better than M-based methods, they still suffer from the problems of IR small-target information loss in the deep networks and inaccurate GT labelling. Thus, the existing IRSTD methods for complex backgrounds have poor robustness and low detection accuracy.

\end{enumerate}

\section{Methodology}
\label{sec:Methodology}
\subsection{Overall Architecture}
As illustrated in \Cref{fig:Overall_Architecture}, the proposed SSHD-Net uses an IR image as the input.
First, we extract the multi-scale features of IR small-target using a high-resolution cross-feature extraction module (HCEM). We then perform dynamic feature enhancement and fusion of the feature maps using a dynamic coordinate fusion module (DCFM). Subsequently, a high-resolution multilevel residual module (HMRM) is applied to generate deep semantic information about the target, and multiple features are further enhanced by the DCFM. Finally, the detection results are output using an adaptive target localization detection head (ATLDH).

This section is organized as follows: In \Cref{subsec:HCEM}, we introduce the high-resolution cross-feature extraction module. \Cref{subsec:DCFM} describes dynamic coordinate fusion module in detail. \Cref{subsec:HMRM} describes the structure and computational process of the high-resolution multilevel residual module. Finally, in \Cref{subsec:ATLDH}, we propose an adaptive target localization detection head.

\subsection{High-Resolution Cross-Feature Extraction Module}
\label{subsec:HCEM}

\begin{figure} 
    \centering
    \includegraphics[width=0.9\linewidth]{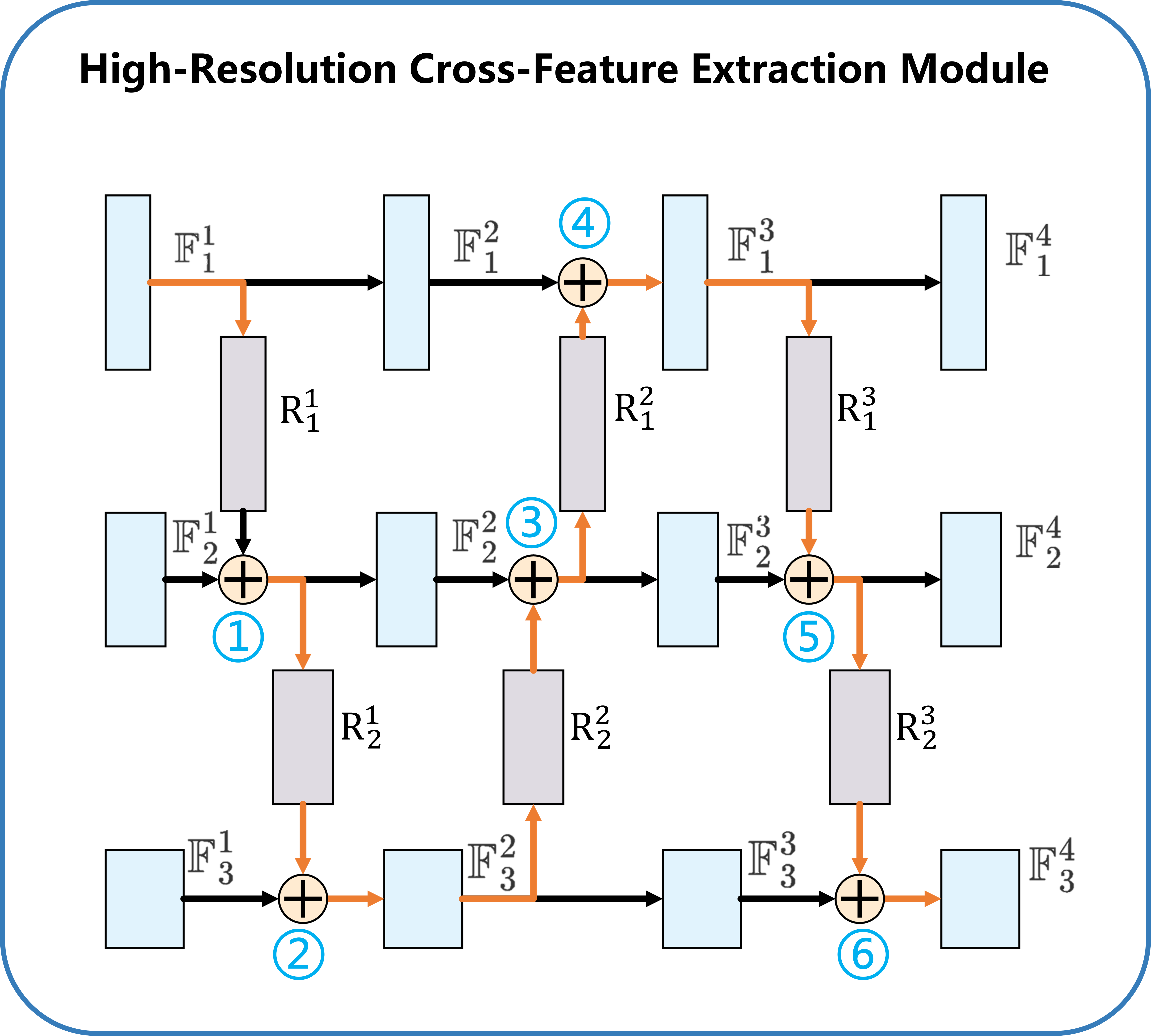}
    \caption{High-resolution cross-feature extraction module (HCEM). During the propagation process, HCEM maintains the parallel propagation of high- and low-resolution features and achieves bi-directional feature interaction through the stepped feature cascade channel (SFCC) as shown by the \textcolor{myorange}{orange lines}, which maintains the balance between network depth and feature resolution. This effectively preserves the deep IR small-target information.}
    \label{fig:HCEM}
\end{figure}

When extracting deep features, existing networks \cite{he2016resnet}\cite{howard2017mobilenets} typically apply techniques such as pooling to downsample the feature map significantly, resulting in the loss of small target information. By contrast, if we reduce the depth of the network, it lacks sufficient target semantic information extraction capability to overcome clutter interference from the complex background. To address these problems, we propose a high-resolution cross-feature extraction module (HCEM) that aims to achieve a delicate balance between the network depth and feature map size.

Inspired by \cite{sun2019hrent}, HCEM maintains the parallel propagation of high- and low-resolution feature layers throughout the process to ensure that the network fully retains the IR small-target information in the deep layer. On this basis, we design the stepped feature cascade channel (SFCC) as shown by the \textcolor{myorange}{orange lines} in \Cref{fig:HCEM} to achieve a progressive bi-directional interaction between feature maps of different scales. During network propagation, high- and low-resolution features are fused step by step through nodes \ding{172} and \ding{173} or \ding{176} and \ding{177} to enrich the contextual information of low-resolution features, as illustrated in \Cref{fig:HCEM}. At nodes \ding{174} and \ding{175}, the global information captured by the low-resolution features is fed back to the high-resolution feature layer step by step through SFCC to enhance the robustness of target localization in complex scenes.

HCEM consists of basic blocks of $I$ rows and $J$ columns for feature extraction. For specific illustration, we select $i (i=0,1,2,... ,I)$ and $j (j=0,1,2,... ,J)$ to describe the structure of HCEM, as illustrated in \Cref{fig:HCEM}. We set ${\mathbb{F}}_{j}^{i} \in {\mathbb{R}}_{}^{{C}_{j}\times {H}_{j}\times {W}_{j}} $ to denote the output of node $\hat{\mathbb{F}}_{j}^{i}$, where $i$ and $j$ denote columns and rows, respectively. When $i = 1$, each node receives the feature map $X \in {\mathbb{R}}_{}^{{C}_{j}\times {H}_{j}\times {W}_{j}} $ from different scales as input. The feature map ${\mathbb{F}}_{j}^{i}$ can be expressed as:

\begin{equation}
\label{eq_fe}
{\mathbb{F}}_{j}^{i} = X +\textup{BN}(\textup{Conv}(\delta (\textup{BN}(\textup{Conv}(X))))),
\end{equation}
where $\textup{Conv}$ denotes convolutional layer whose kernel size is $ C \times C \times 3 \times 3$. $\delta$ denotes the rectified linear unit (ReLU). And BN denotes batch normalisation.
 
To achieve progressive bi-directional interaction of multi-scale feature maps, the SFCC module connects different layers sequentially during feature map propagation. When $i > 1$ and even, the input ${{\mathbb{I}}_{j}^{i}} \in {\mathbb{R}}_{}^{{C}_{j}\times {H}_{j}\times {W}_{j}} $ of each node $\hat{\mathbb{F}}_{j}^{i}$ can be expressed as: 

\begin{equation}
\label{eq_input_even}
{{\mathbb{I}}_{j }^{i}} = {\mathbb{F}}_{j}^{i-1} + {R}_{j-1}^{i-1},
\end{equation}
where ${R}_{j}^{i}$ denotes the downsampling and channel alignment. When $i>1$ and odd, the ${{\mathbb{I}}_{j}^{i}}$ can be expressed as:
\begin{equation}
\label{eq_input_odd}
{{\mathbb{I}}_{j }^{i}} = {\mathbb{F}}_{j}^{i-1} + {R}_{j}^{i-1},
\end{equation}
In the above \Cref{eq_input_even}, \Cref{eq_input_odd}, the subscript $j$ of ${R}{j}^{i}$ belongs to the set [1, 2]; for other values of $j$, ${R}_{j}^{i}$ is defined as 0.

\subsection{Dynamic Coordinate Fusion Module}
\label{subsec:DCFM}

\begin{figure*}[!t] 
    \centering
    \includegraphics[width=0.9\textwidth]{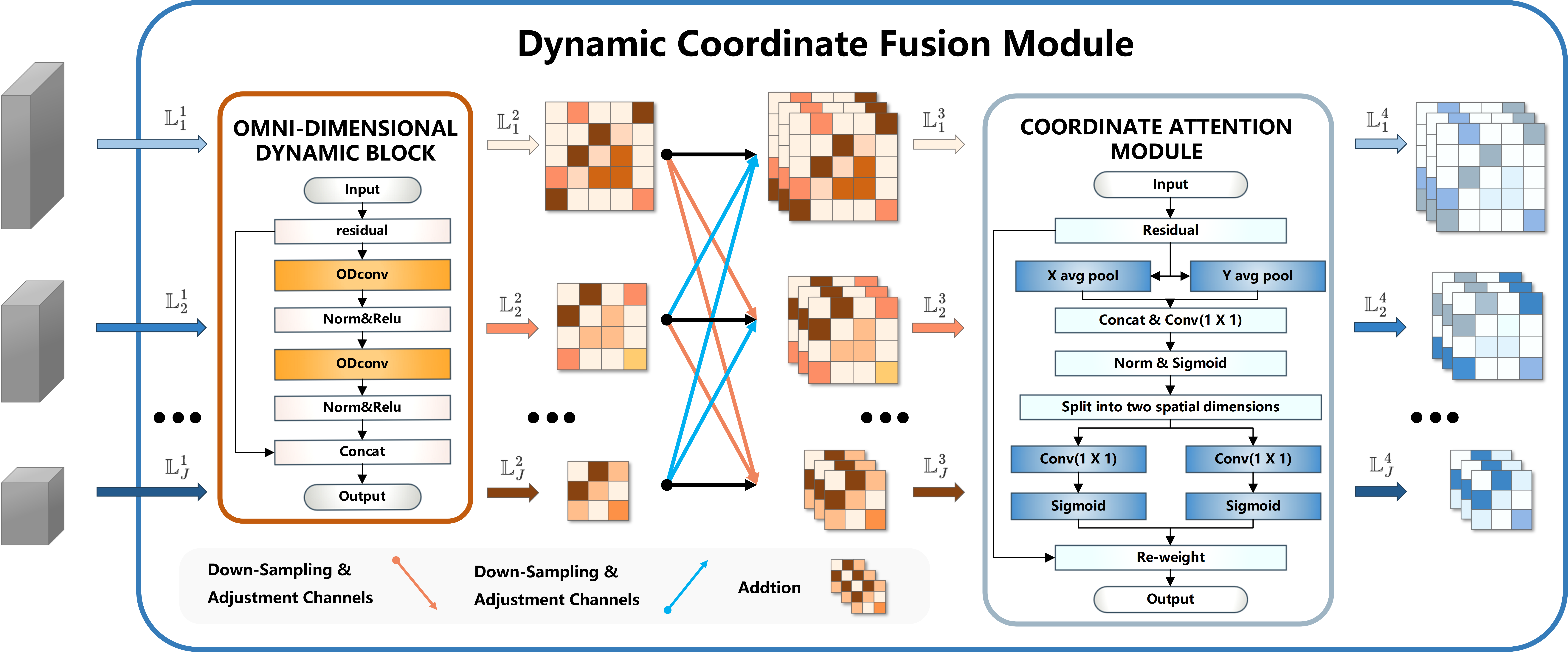}
    \caption{Dynamic coordinate fusion module (DCFM). Firstly, the features ${\mathbb{L}}_{j}^{1}$ are processed through the full dimensional dynamic block to get ${\mathbb{L}}_{j}^{2}$, and then the features of different layers are fused to get ${\mathbb{L}}_{j}^{3}$. Finally, the features are passed through the CA module to get the output features ${\mathbb{L}}_{j}^{4}$.}
    \label{fig:DCFM}
\end{figure*}

\begin{figure} 
    \centering
    \includegraphics[width=0.9\linewidth]{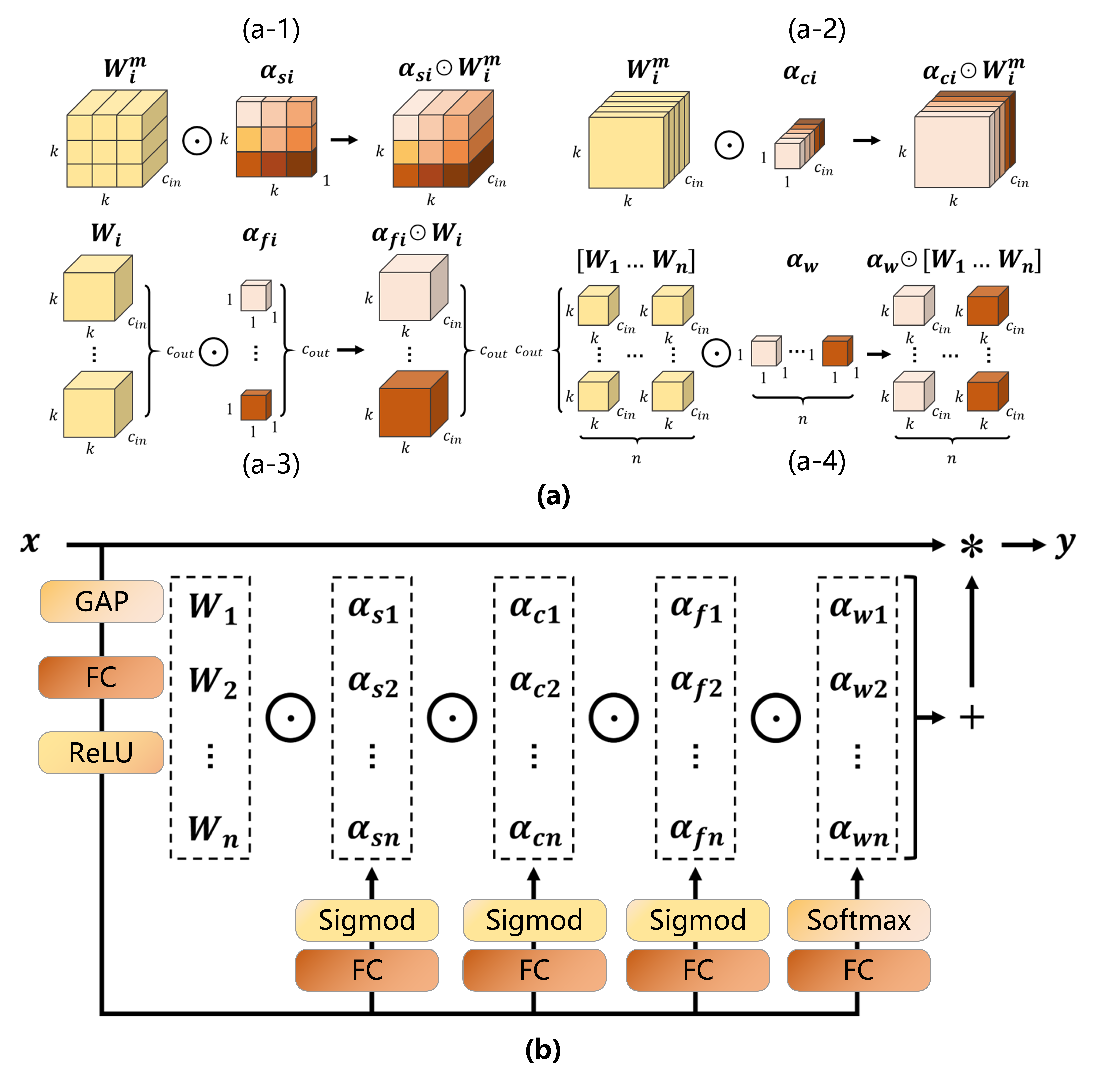}
    \caption{(a) Illustration of multiplying four types of attentions in ODConv to convolutional kernels progressively. (a-1) Location-wise multiplication operations along the spatial dimension, (a-2) channelwise multiplication operations along the input channel dimension, (a-3) filter-wise multiplication operations along the output channel dimension, and (a-4) kernel-wise multiplication operations along the kernel dimension of the convolutional kernel space. (b) ODconv. ODConv leverages a novel multi-dimensional attention mechanism to compute four types of attentions ${\alpha }_{si}$, ${\alpha }_{ci}$, ${\alpha }_{fi}$ and ${\alpha }_{wi}$ for ${W}_{i}$ along all four dimensions of the kernel space in a parallel manner. Their formulations and implementations are clarified in the \Cref{subsec:DCFM}.}
    \label{fig:ODBlock}
\end{figure}

In a network structure with multi-resolution parallel propagation, feature layers at the same depth are capable of exhibiting multiple semantic features.The high-resolution layer can retain exhaustive target detail information, whereas the low-resolution layer focuses on providing extensive global information. To efficiently utilize this contextual information and achieve more accurate target localization, we design the Dynamic Coordinate Fusion Module (DCFM), as illustrated in \Cref{fig:DCFM}. It consists of an omni-dimensional dynamic convolution block (ODBlock), feature fusion module, and coordinate attention (CA) module.

Specifically, ODBlock focuses on essential information and suppresses complex background interference by adaptively adjusting the size, spatial dimensions, and input and output channels of the convolution kernel. Thus, it can effectively capture IR small-target information from global to local details. The ODBlock can be expressed as:

\begin{equation}
\label{ODBlock}
{\mathbb{L}}_{j}^{2} = X +\textup{BN}(\textup{ODConv}(\delta (\textup{BN}(\textup{ODConv}({\mathbb{F}}_{j}^{4}))))),
\end{equation}
where ${\mathbb{L}}_{j}^{2}$ denotes the output feature map of ${\mathbb{F}}_{j}^{4}$ after the ODBlock. The $\textup{ODConv}$ denotes omni-dimensional dynamic convolution (ODconv)\cite{li2022odconv}. And we select $j (j=0,1,2,... ,J)$ to describe the structure of the DCFM.

Specifically, the ODconv can be expressed as:

\begin{equation}
\label{ODconv}
     \begin{aligned}
      y = &({\alpha }_{w1}\odot {\alpha }_{f1}\odot {\alpha }_{c1}\odot {\alpha }_{s1}\odot {W}_{1}+...\\
       &+{\alpha }_{wn}\odot {\alpha }_{fn}\odot {\alpha }_{cn}\odot {\alpha }_{sn}\odot {W}_{n})\times x,
     \end{aligned}
\end{equation}
where $x \in {\mathbb{R}}_{}^{{C}_{in}\times {H}\times {W}} $ and $y \in {\mathbb{R}}_{}^{{C}_{out}\times {H}\times {W}} $ denote the input features and the output features(having ${C}_{in}/{C}_{out}$ channels with the height $H$ and the width $W$), respectively;${W}_{i}$ denotes the ${i}^{th}$ convolutional kernel consisting of ${C}_{out}$ filters ${W}_{i}^{m}\in {\mathbb{R}}_{}^{{C}_{in}\times {K}\times {K}}, m=1,..., {C}_{out} $; ${\alpha }_{wi} \in \mathbb{R}$ is the attention scalar for weighting ${W}_{i}$; ${\alpha }_{si} \in {\mathbb{R}}^{K \times K}$, ${\alpha }_{ci} \in {\mathbb{R}}^{{C}_{in}}$ and ${\alpha }_{fi} \in {\mathbb{R}}^{{C}_{out}}$ denote three attentions which are computed along the spatial dimension, the input and the output channel dimensions of the kernel space for the convolutional kernel ${W}_{i}$
, respectively; $\odot $ denotes the  multiplication operations along different dimensions of the kernel space; $*$ denotes the convolution operation. Here, ${\alpha }_{wi}$, ${\alpha }_{si}$, ${\alpha }_{fi}$ and ${\alpha }_{ci}$ are computed with a multi-head attention module ${\pi}_{i}(x)$.

Subsequently, the multi-scale feature fusion module is applied to achieve feature fusion using the width-height and channel alignment of the $J$ feature maps. Finally, to refine the spatial distribution of the target in the feature maps and enhance the ability to perceive the precise location of the target, we apply the coordinate attention (CA) \cite{hou2021coordinateAtt} module to enhance the fused feature maps and ensure that the computational resources are focused on allocating to the key regions. The basic process of the CA module is as follows. 

First, we encode each channel of the input feature map ${\mathbb{L}}_{j}^{3}$ using two spatially scoped pooling kernels $(H, 1)$ or $(1, W)$ along the horizontal and vertical coordinates, respectively. The formulae are as follows:
\begin{equation}
\label{CA1_h}
{Z}_{c}^{h}(h)=\frac{1}{W}\displaystyle\sum_{0\leqslant t\leqslant W}{{\mathbb{L}}_{j}^{3}}_{c}(h,t),
\end{equation}
where ${Z}_{c}^{h}(h)$ denotes the output of the ${c}^{th}$ channel at height $h$.
\begin{equation}
\label{CA1_w}
{Z}_{c}^{w}(w)=\frac{1}{W}\displaystyle\sum_{0\leqslant r\leqslant H}{{\mathbb{L}}_{j}^{3}}_{c}(r,w),
\end{equation}
similarly, ${Z}_{c}^{w}(w)$ denotes the output of the ${c}^{th}$ channel at width $w$.

Then, in order to generate coordinate attention, the target position information is fully captured. We perform the following calculations on the feature maps generated in \Cref{CA1_h} and \Cref{CA1_w}.

\begin{equation}
\label{CA2_f}
f = \delta ({\textup{F}}_{1}([{z}^{h},{z}^{w}])),
\end{equation}
where ${\textup{F}}_{1}(x)$ denotes a shared $1\times1$ convolutional transformation function.$[\bullet, \bullet ]$ denotes the concatenation operation along the spatial dimension, $\delta$ is a non-linear activation function and $f\in {\mathbb{R}}^{C/r\times (H+W)}$ is the intermediate feature map that encodes spatial information in both the horizontal direction and the vertical direction.Here, $r$ denotes the reduction ratio for controlling the block size.

Subsequently, we split $f$ into two separate tensors ${f}^{h}\in {\mathbb{R}}^{C/r\times (H)}$ and ${f}^{w}\in {\mathbb{R}}^{C/r\times (W)}$ along the spatial dimension and perform the following calculations.
\begin{equation}
\label{CA3_gh}
{g}^{h}=\sigma ({\textup{F}}_{h}({f}^{h})),
\end{equation}

\begin{equation}
\label{CA3_gw}
{g}^{w}=\sigma ({\textup{F}}_{w}({f}^{w})),
\end{equation}
where function ${\textup{F}}_{h}(x)$ and ${\textup{F}}_{w}(x)$ have the same action as function ${F}_{1}(x)$.Here, $\sigma$ is the sigmoid function.

 Finally, the input ${\mathbb{L}}_{j}^{3}$ is weighted and encoded with spatial information of different dimensions as attention weights, and the expression of the output ${\mathbb{L}}_{j}^{4}$ is:

 \begin{equation}
\label{CA4_L}
{\mathbb{L}}_{j}^{4}(t,r)={\mathbb{L}}_{j}^{3}(t,r)\times{g}_{c}^{h}(t)\times {g}_{c}^{w}(r).
\end{equation}

\subsection{High-resolution Multilevel Residual Module}
\label{subsec:HMRM}
To extract deep IR small-target information effectively , we design a high-resolution multilevel residual module (HMRM). This network uses a nested jump-connection structure, which helps extract deeper semantic information. The calculation process is as follows:
\begin{equation}
\label{eq_h1}
{\mathbb{H}}_{j}^{1} = {\mathbb{L}}_{j}^{4} +\textup{BN}(\textup{Conv}(\delta (\textup{BN}(\textup{Conv}({\mathbb{L}}_{j}^{4}))))),
\end{equation}
where ${\mathbb{H}}_{j}^{1} \in {\mathbb{R}}_{}^{{C}_{j}\times {H}_{j}\times {W}_{j}} $ denotes output feature map of ${\mathbb{L}}_{j}^{4}$ through a basic residual block.

\begin{equation}
\label{eq_h2}
{\mathbb{H}}_{j}^{2} = {\mathbb{L}}_{j}^{4} + \textup{BN}(\textup{Conv}(\delta (\textup{BN}(\textup{Conv}({\mathbb{L}}_{j}^{4}+{\mathbb{H}}_{j}^{1}))))),
\end{equation}
where ${\mathbb{H}}_{j}^{2}$ denotes the final output of the HMRM. Without loss of generality, we set $j (j=0,1,2,... ,J)$ to describe the structure of HMRM. In addition, all the symbols in \Cref{eq_h2} are the same as those defined in \Cref{eq_fe}.

\subsection{Adaptive Target Localization Detection Head}
\label{subsec:ATLDH}


\begin{figure*} 
    \centering
    \includegraphics[width=0.75\linewidth]{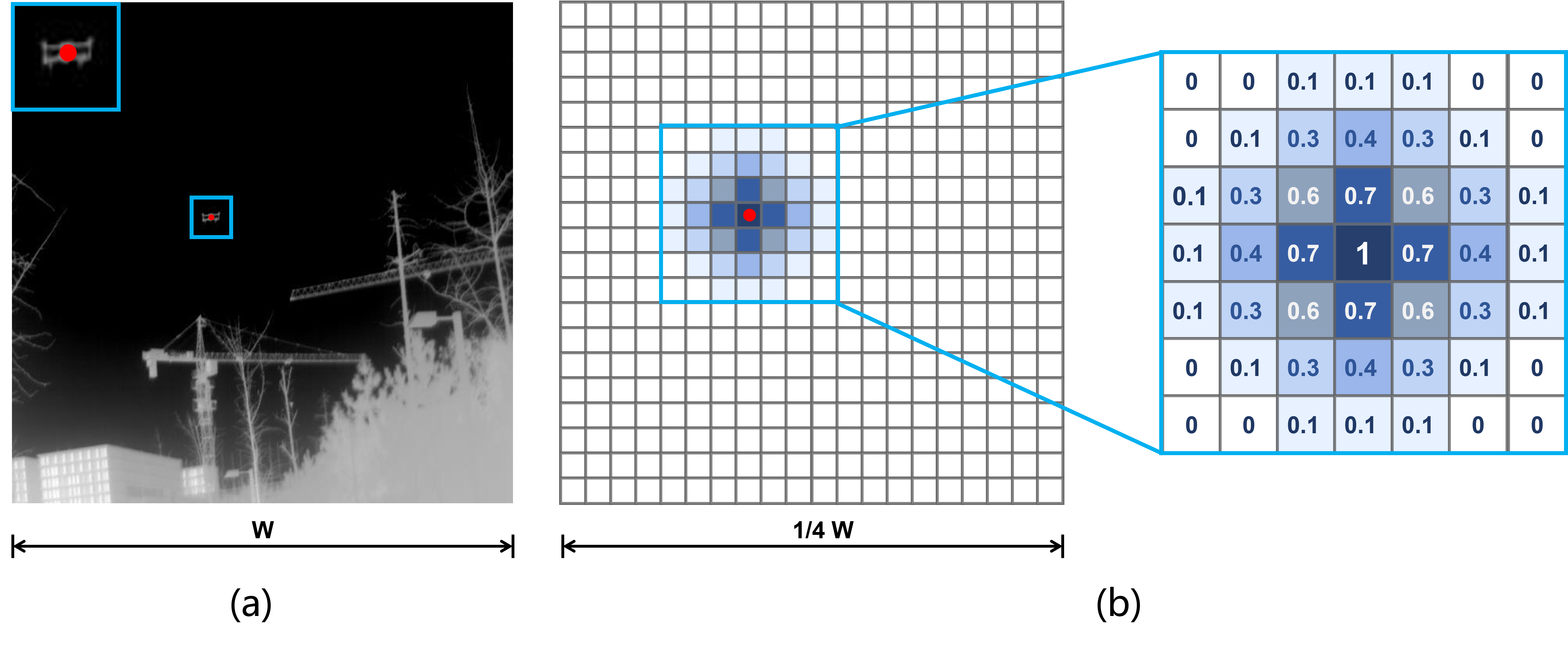}
    \caption{Generation of Gaussian distribution heatmap. (a) Single-point labelling. (b) Heatmap of Gaussian probability distribution. as illustrated in the figure on the right, the Gaussian distribution heatmap radiates in all directions centred on a Single-point labelling. The numbers in the figure represent the probability value of the point being a the target, which ranges from [0,1].}
    \label{fig:Heatmap}
\end{figure*}

\begin{figure} 
    \centering
    \includegraphics[width=0.85\linewidth]{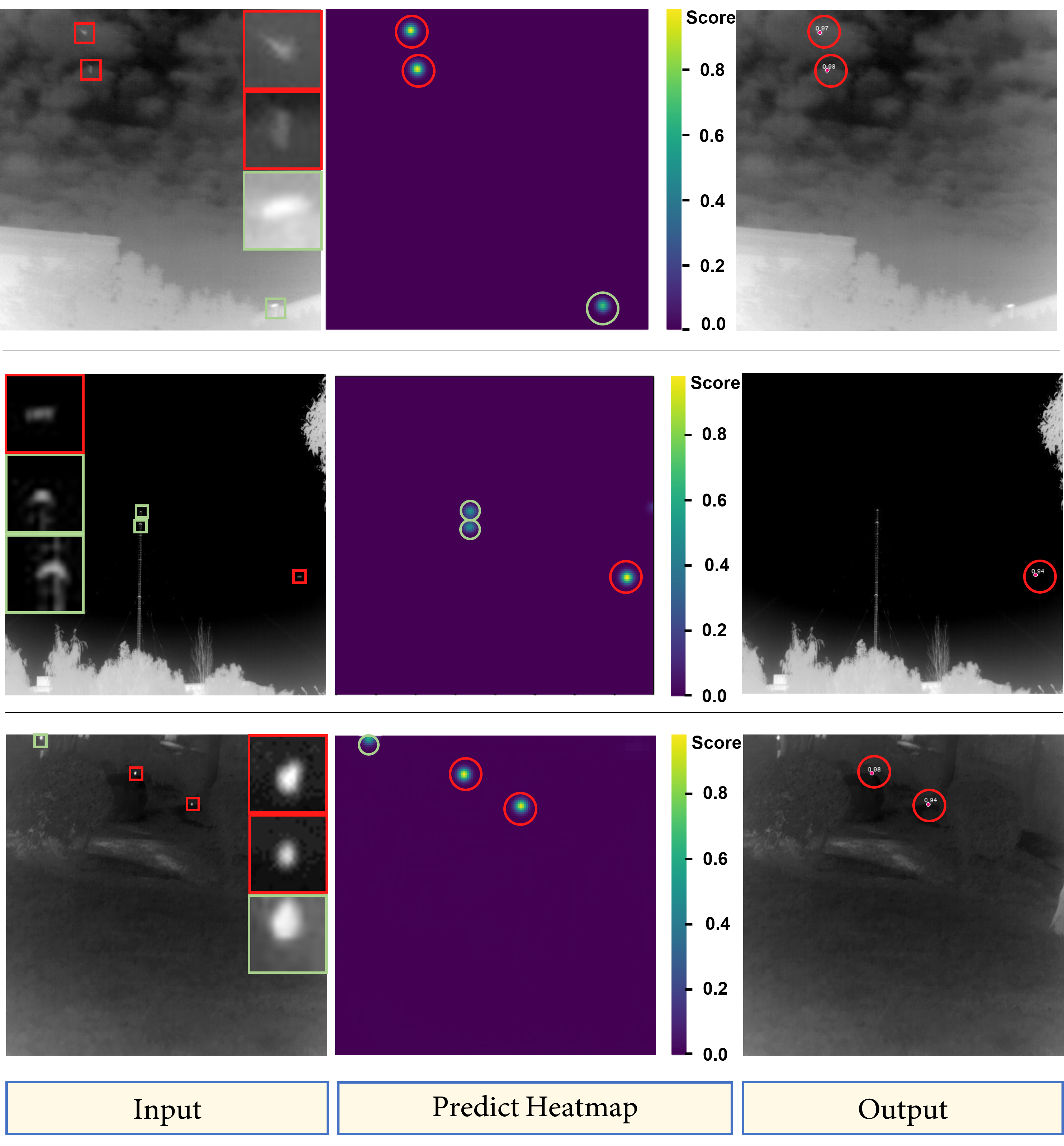}
    \caption{Adaptive target localization detection head (ATLDH) visualisation result figure. In the input image, the \textcolor{myred}{red box} identifies the target and the \textcolor{mygreen}{green box} identifies the background interferences that are similar to the target. In the predict heatmap and the output image, \textcolor{myred}{red circles} indicate correct detection and \textcolor{mygreen}{green circles} indicate false detection. as illustrated in the figure, ATLDH can effectively suppress the background interference and significantly improve the precision of the method.}
    \label{fig:ATLDH}
\end{figure}

Owing to the influence of the thermal-effect of infrared radiation, IR small-target usually exhibit blurred target edges, making it difficult to clearly distinguish the boundary between the target and background. 
In detection-based networks, small perturbations in the bounding box can significantly affect the intersection-incorporation ratio (IOU) metrics, making the regression of the target box more challenging. Moreover, unclear bounding box labelling may lead to incorrect positive and negative sample assignments, thus degrading the performance of the detector.
In segmentation-based networks, uncertainty mask labelling can lead to incorrect treatment of the background as a target,  thereby increasing the difficulty of classifying semantic information. 

To address these issues, we designed an adaptive target localization detection head (ATLDH) capable of detecting IR small-target using point-to-point target regression. This method allows the network to focus on the central region of the target, thereby effectively reducing the uncertainty caused by boundary blurring. We describe ATLDH in detail from the following three aspects, Gaussian heatmap, loss function and adaptive non-maximal suppression.

\begin{enumerate}[wide] 
\item{\textit{\textbf{Gaussian Heatmap}}:} We start by labelling the target with a single point(i.e., a point in the central region of the target mask under a Gaussian distribution is selected as the single-point ground truth (SPGT)), and a 2D Gaussian probability distribution ground truth (GT) heatmap centered on the point and diverging in all directions is generated, as illustrated in \Cref{fig:Heatmap}. Gaussian distribution can be expressed as follows: 

\begin{equation}
\label{eq_Gauss}
\textup{G}(x, y) = e^{-\frac{x^2 + y^2}{2{\sigma}_{G}^2}},
\end{equation}
where $x$ and $y$ denote the distance (horizontal and vertical, respectively) from the core, and ${\sigma}_{\textup{G}}$ is the standard deviation of the Gaussian distribution, which controls the width of the distribution. The probability score at position $({i}_{h},{j}_{h})$ in the GT heatmap can be expressed as follows:
\begin{equation}
\label{eq_Heatmap}
\textup{heatmap}({i}_{h},{j}_{h}) = \textup{G}({i}_{h}-k{x}_{k},{j}_{h}-k{y}_{k}),
\end{equation}
where $\textup{G}(x, y)$ denotes the Gaussian kernel function defined in \Cref{eq_Gauss}, $k$ denotes the index of the target, and $k{x}_{k}$, $k{y}_{k}$ denote the coordinate position of the ${k}^{th}$ target point.

\item{\textit{\textbf{Loss Function}}:} IR images are fed into SSHD-Net to obtain the predicted heatmap. As \cite{sun2019hrent}, we use the mean squared error (MSE) for the GT and predicted heatmaps to guide the training of the network. MSE can be expressed as follows:
\begin{equation}
\label{eq_MSE}
  \begin{aligned}
    \textup{MSE} = \frac{1}{n} \sum_{{i}_{h}=1}^{{I}_{h}} \sum_{{j}_{h}=1}^{{J}_{h}} \Bigl(&\hat{\textup{heatmap}}(i_{h},j_{h}) - \\
    &\textup{heatmap}(i_{h},j_{h})\Bigr)^{2},
  \end{aligned}
\end{equation}

where $\hat{\textup{heatmap}}({i}_{h},{j}_{h})$ denotes the value of the predicted heatmap at coordinate $({i}_{h},{j}_{h})$, ${I}_{h}$ and ${J}_{h}$ denote the width and height of the heatmap, respectively. The other variables are as defined in \Cref{eq_Heatmap}.

\item{\textit{\textbf{Adaptive Non-Maximal Suppression}}:} It is difficult to effectively suppress background interference by relying only on the probability score from the predicted heatmap, as illustrated in \Cref{fig:ATLDH}. There are often numerous background interferences that are similar to those of the target in a real scene. Therefore, we design an adaptive non-maximal suppression (ANMS) method to enhance interference suppression. The method adaptively adjusts the background suppression threshold according to the target confidence. The specific implementation steps are as follows: First, we use a $3\times3$ pooling layer to retain the local maximum matrix $\textup{Maxm}$, which can be expressed as: 

\begin{equation}
    \label{eq_Maxm}
    \textup{Maxm} = \textup{MaxPool}(\textup{heatmap}({i}_{h},{j}_{h})),
\end{equation}
where $\textup{MaxPool}$ denotes the maximum pooling layer with the kernel size of $3\times3$. Then, we compare the values in the heatmap with $\textup{Maxm}$, and if they are equal, the values are retained; otherwise, they are set to 0; the formula for the transformed matrix ${\textup{heatmap}}^{\prime}({i}_{h},{j}_{h})$ can be expressed as:
\begin{equation}
    \label{eq_heatmap'}
    \begin{aligned}
    {\textup{heatmap}}^{\prime}({i}_{h},{j}_{h}) = &{\textup{heatmap}}({i}_{h},{j}_{h})\times\\
    &(\textup{heatmap}({i}_{h},{j}_{h})==\textup{Maxm}).
    \end{aligned}
\end{equation}

Subsequently, we select the target point ${\textup{heatmap}}_{max}^{\prime}({i}_{h},{j}_{h})$ with the highest confidence level from the ${\textup{heatmap}}^{\prime}$ as the benchmark point. And calculate the confidence difference between other target points and the benchmark point. If the difference is within a certain range, the point is regarded as a valid target point; beyond that range, it is regarded as the background and eliminated. The formula can be expressed as follows: 

\begin{equation}
    \label{eq_atldh_output}
    \begin{aligned}
    &if|{\textup{heatmap}}^{\prime}({i}_{h},{j}_{h})-{\textup{heatmap}}_{max}^{\prime}({i}_{h},{j}_{h})|> \\
    &\lambda \times \textup{max}({\textup{heatmap}}^{\prime}({i}_{h},{j}_{h}),{\textup{heatmap}}_{max}^{\prime}({i}_{h},{j}_{h})).
    \end{aligned}
\end{equation}
If the above equation holds, the point is suppressed as background; conversely, it is selected as the target point.

\end{enumerate}

\section{Experiment}
\label{sec:Experiment}
\subsection{Experimental Settings}
\label{subsec: Experimental Settings}

For the experiment, we select two public IR small-target datasets: NUDT-SIRST \cite{li2022DNA} and IRSTD-1k \cite{zhang2022isnet_msaff16}. First, SPGT are performed annotation on both datasets. We conduct experiments using different methods with bounding boxes, semantic segmentation masks, and single-point annotations. To provide a fair experimental environment, we re-divide each dataset into training, validation and test sets in a ratio of 6:2:2.  During the experiment, the M-based methods are tested on a Lenovo computer equipped with an Intel Core I7-11800H CPU, and the DL-based methods are evaluated on a Lenovo computer equipped with an Intel Core I7-11800H CPU and an Nvidia GeForce RTX 3090 GPU. In addition, the evaluation metrics and comparative SOTA models are described in detail in \Cref{subsec: Experimental Settings} 1) and \Cref{subsec: Experimental Settings} 2), respectively.

\begin{enumerate}[wide] 
\item{\textit{\textbf{Evaluation Metrics}}:} Due to the properties of IR imaging and the need for early warning by detection systems, IR small-target are usually very small and contain only a few pixels. And in the process of practical application, accurate target location is the core evaluation criterion of IRSTD. Therefore, we choose the target-level evaluation metrics, including precision (Pre), recall (Rec) and F1-measure (F1). Each metric is defined as follows:

Precision measures the proportion of true positive predictions out of all positive predictions made, including both true positives and false positives. The formula can be expressed as:
\begin{equation}
\label{eq_pre}
\textup{Precision} = \frac{\textup{TP}}{\textup{TP}+\textup{FP}},
\end{equation}
where TP represents the true positives (i.e., the number of positive samples correctly identified), and FP represents the false positives (i.e., the number of negative samples incorrectly identified as positive).

Recall assesses the proportion of true positive instances correctly identified by the model out of all actual positive instances, including both true positives and false negatives. The formula can be expressed as:
\begin{equation}
\label{eq_rec}
\textup{Recall} = \frac{\textup{TP}}{\textup{TP}+\textup{FN}},
\end{equation}
where TP is as defined above, and FN represents the false negatives (i.e., the number of positive samples not recognized by the model).

The F1-measure refers to the reconciled average of Pre and Rec, and is used to comprehensively assess the balancing effect of Pre and Rec. The formula can be expressed as:
\begin{equation}
\label{eq_f1}
\textup{F1} = 2\times\frac{\textup{Precision}\times\textup{Recall}}{\textup{Precision}+\textup{Recall}}.
\end{equation}

To compare the modules performance for different tasks fairly, we uniformly use the above evaluation metrics in our experiments. For segmentation-based models, we cluster pixels belonging to the same target together using eight-connected neighbourhood clustering\cite{li2022DNA}, and select the center of the target mask under a Gaussian distribution as the predicted target. When the Euclidean distance between the prediction result and the SPGT is within five pixels, it is considered TP; beyond that, it is considered FP. Similarly for the detection-based module, we evaluate the Euclidean distance between the centroids of the predicted boxes and the SPGT. Distances within five pixels are considered TP, whereas those beyond this range are considered FP.

\item{\textit{\textbf{Comparison with the SOTA Methods}}:}
To demonstrate the superiority of the propsed method, we compare SSHD-Net with a variety of SOTA methods. These methods include the M-based methods FKRW\cite{qin2019FKRW_qiu120}, IPI\cite{gao2013IPI_efl8}, MPCM\cite{wei2016MPCM_qiu110}, and PSTNN\cite{zhang2019PSTNN_efl24}, as well as the DL-based methods MDvsFA-cGAN\cite{wang2019MDvsFA_efl10}, ACM-Net\cite{dai2021acm_efl11}, AGPC-Net\cite{zhang2023agpc}, DNA-Net\cite{li2022DNA}, UIU-Net\cite{wu2022uiu} and EFL-Net\cite{yang2024eflnet}. To ensure a fair comparison, all DL-based methods are trained on the same dataset as SSHD-Net, with the parameter settings recommended in the original paper. All M-based methods are test with according to the parameter settings listed in \Cref{tab:Hyperparmeter Settings Of Model-Based Methods}.
\end{enumerate}

\subsection{Quantitative Results}

\begin{table*}[]
\caption{ Detailed Hyperparmeter Settings Of Model-Based Methods For Comparison\label{tab:Hyperparmeter Settings Of Model-Based Methods}}
\centering
\renewcommand{\arraystretch}{1.5} 
\setlength{\tabcolsep}{12mm}
\begin{tabular}{c|c}
\thickhline
Methods & Hyper-parameter settings                  \\ \hline
MPCM\cite{wei2016MPCM_qiu110}    & $N$ = $1, 3,..., 9$, threshold factor: $k=3 $     \\
FKRW\cite{qin2019FKRW_qiu120}    & $K=5$, $p=6$, $\beta = 200$, window size: $11 \times 11 $      \\
IPI\cite{gao2013IPI_efl8}     & Patch size: $50 \times 50$, stride: $20$,  $\lambda =L/{min(m,n)}^{1/2}$, $L = 2.5$, $\epsilon = {10}^{-7}$              \\ 
PSTNN\cite{zhang2019PSTNN_efl24}   & Patch size: $40 \times 40$, Step: $40$, $\lambda =1/ \sqrt{min({n}_{1},{n}_{2})*{n}_{3} }$, $\beta =0.01$, $\mu =200$ \\
\thickhline
\end{tabular}
\end{table*}

\begin{table*}[]
\caption{ Comparison With SOTA Methods On  NUDT-SIRST And IRSTD-1K In Precision, Recall, AND F1-measure\label{tab:comparison with sota}}
\centering
\renewcommand{\arraystretch}{1.5} 
\setlength{\tabcolsep}{5mm}
\begin{tabular}{cccccccc}
\thickhline

\multicolumn{1}{c}{}                           &                                & \multicolumn{3}{c}{\textbf{NUDT-SIRST}}                                                                               & \multicolumn{3}{c}{{\color[HTML]{393939} \textbf{IRSTD-1k}}}                                                          \\ \cline{3-8} 
\multicolumn{1}{c}{\multirow{-2}{*}{\textbf{Category}}} & \multirow{-2}{*}{\textbf{Methods}}       & {\color[HTML]{393939} \textbf{Pre}}   & {\color[HTML]{393939} \textbf{Rec}}   & {\color[HTML]{393939} \textbf{F1}}    & {\color[HTML]{393939} \textbf{Pre}}   & {\color[HTML]{393939} \textbf{Rec}}   & {\color[HTML]{393939} \textbf{F1}}    \\ \hline
                                               & {\color[HTML]{393939} MPCM\cite{wei2016MPCM_qiu110}}    & {\color[HTML]{393939} 12.60}          & {\color[HTML]{393939} 89.46}          & {\color[HTML]{393939} 22.09}          & {\color[HTML]{393939} 21.85}          & {\color[HTML]{393939} 84.51}          & {\color[HTML]{393939} 34.72}          \\
                                               & {\color[HTML]{393939} FKRW\cite{qin2019FKRW_qiu120}}    & {\color[HTML]{393939} 38.45}          & {\color[HTML]{393939} 67.61}          & {\color[HTML]{393939} 49.02}          & {\color[HTML]{393939} 33.22}          & {\color[HTML]{393939} 63.64}          & {\color[HTML]{393939} 43.65}          \\  
                                               & {\color[HTML]{393939} IPI\cite{gao2013IPI_efl8}}     & {\color[HTML]{393939} 35.28}          & {\color[HTML]{393939} 80.20}          & {\color[HTML]{393939} 49.00}          & {\color[HTML]{393939} 44.76}          & {\color[HTML]{393939} 65.31}          & {\color[HTML]{393939} 53.11}          \\
\multirow{-4}{*}{\textbf{Model-Based}}          & {\color[HTML]{393939} PSTNN\cite{zhang2019PSTNN_efl24}}   & {\color[HTML]{393939} 39.44}          & {\color[HTML]{393939} 70.88}          & {\color[HTML]{393939} 50.68}          & {\color[HTML]{393939} 44.53}          & {\color[HTML]{393939} 60.20}          & {\color[HTML]{393939} 51.19}          \\       \hline
                                               & {\color[HTML]{393939} MDvsFA-cGan\cite{wang2019MDvsFA_efl10}}  & {\color[HTML]{393939} 79.05}          & {\color[HTML]{393939} 78.42}          & {\color[HTML]{393939} 78.73}          & {\color[HTML]{393939} 83.70}          & {\color[HTML]{393939} 64.85}          & {\color[HTML]{393939} 73.08}          \\
                                               & {\color[HTML]{393939} AGPC-Net\cite{zhang2023agpc}} & {\color[HTML]{393939} 95.95}          & {\color[HTML]{393939} 97.68}          & {\color[HTML]{393939} 96.81}          & {\color[HTML]{393939} 77.48}          & {\color[HTML]{393939} 87.46}          & {\color[HTML]{393939} 82.17}          \\
                                               & {\color[HTML]{393939} ACM-Net\cite{dai2021acm_efl11}}     & {\color[HTML]{393939} 89.38}          & {\color[HTML]{393939} 68.50}          & {\color[HTML]{393939} 77.65}          & {\color[HTML]{393939} 88.03}          & {\color[HTML]{393939} 85.32}          & {\color[HTML]{393939} 86.66}          \\
                                               & {\color[HTML]{393939} DNA-Net\cite{li2022DNA}}  & {\color[HTML]{393939} 98.16}          & {\color[HTML]{393939} 96.14}          & {\color[HTML]{393939} 97.14}          & {\color[HTML]{393939} 82.96}          & {\color[HTML]{393939} 88.05}          & {\color[HTML]{393939} 85.43}          \\
                                               & {\color[HTML]{393939} UIU-Net\cite{wu2022uiu}}  & {\color[HTML]{393939} 91.97}          & {\color[HTML]{393939} 97.93}          & {\color[HTML]{393939} 94.86}          & {\color[HTML]{393939} 85.91}          & {\color[HTML]{393939} 87.37}          & {\color[HTML]{393939} 86.63}          \\
                                               & {\color[HTML]{393939} EFL-Net\cite{yang2024eflnet}}  & {\color[HTML]{393939} 94.33}          & {\color[HTML]{393939} 96.32}          & {\color[HTML]{393939} 95.31}          & {\color[HTML]{393939} 91.44}          & {\color[HTML]{393939} 91.13}          & {\color[HTML]{393939} 91.28}          \\
\multirow{-7}{*}{\textbf{Deep-Learning-Based}}          & \textbf{Ours}                  & {\color[HTML]{393939} \textbf{99.73}} & {\color[HTML]{393939} \textbf{98.42}} & {\color[HTML]{393939} \textbf{99.07}} & {\color[HTML]{393939} \textbf{96.42}} & {\color[HTML]{393939} \textbf{92.76}} & {\color[HTML]{393939} \textbf{94.55}} \\ \thickhline
\end{tabular}
\end{table*}

\Cref{tab:comparison with sota} presents the results of the comparison of the evaluation metrics between the 10 existing methods and SSHD-Net on the NUDT-SIRST\cite{li2022DNA} and IRSTD-1k\cite{zhang2022isnet_msaff16} datasets, with the maximum values in each column highlighted in bold.
The results demonstrate that SSHD-Net achieves the highest scores on all evaluation metrics on both datasets, which verifies that the proposed method has significant advantages over existing SOTA  methods. 

Compared with M-based methods \cite{qin2019FKRW_qiu120}\cite{gao2013IPI_efl8}\cite{wei2016MPCM_qiu110}\cite{zhang2019PSTNN_efl24}, our method performs well on datasets that contain complex scenes (e.g., buildings, jungles, and mountains) with considerable background clutter and noise, low contrast, and small target size. 
In contrast, M-based methods are prone to a large number of false positives in a given scenario, owing to the limitations of manual design. As shown in \Cref{tab:comparison with sota}, although the IPI method performs relatively consistently on the two datasets, the Pre on the NUDT-SIRST and IRSTD-1k datasets is only 35.27$\%$ and 44.76$\%$, respectively, reducing F1 to 49$\%$ and 53.11$\%$, respectively. This is significantly lower than that of the DL-based method.

Compared with DL-based methods \cite{wang2019MDvsFA_efl10}\cite{dai2021acm_efl11}\cite{zhang2023agpc}\cite{li2022DNA}\cite{wu2022uiu} \cite{yang2024eflnet} , SSHD-Net maintains the highest scores on three evaluation metrics: Pre, Rec and F1. On the NUDT-SIRST and IRSTD-1k datasets, SSHD-Net outperforms DNA-Net F1 by 1.93$\%$ and 9.12$\%$, respectively. This is mainly attributed to the fact that SSHD-Net not only maintains the semantic information of IR small-target in the deep layer, but also dynamically enhances the spatial information of the targets. In addition, the F1 of our method on these two datasets are 3.76$\%$ and 3.27$\%$ higher than those of EFL-Net, respectively. The results show that SSHD-Net improves precision and robustness in complex scenarios by focusing the network on the core region of the target through point-to-point regression.

\subsection{Visual Results}

\begin{figure*}[!t]
    \centering
    \includegraphics[width=16cm]{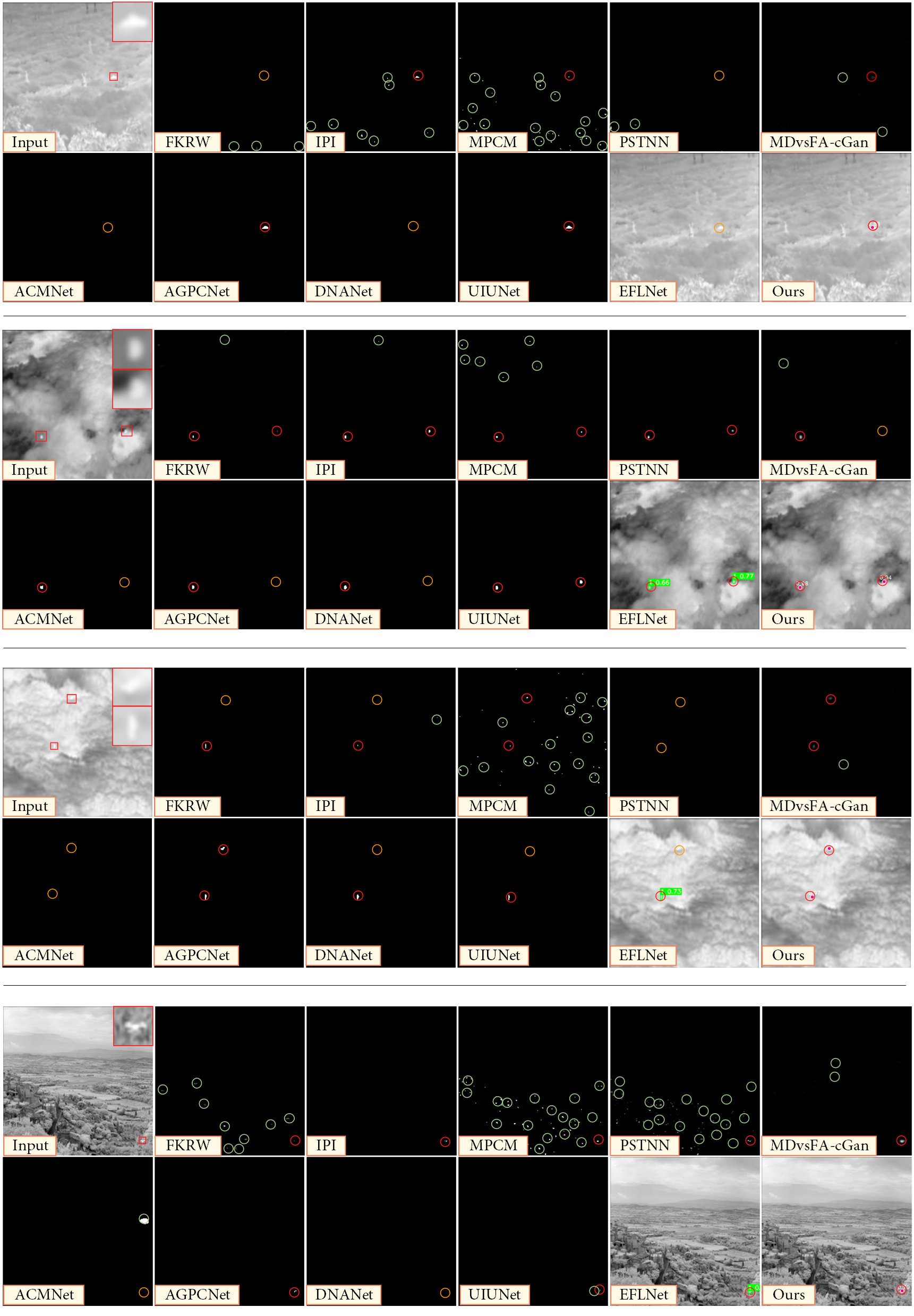}
    \caption{Visual results of different methods on the NUDT-SIRST dataset. In the input images, real targets are indicated by the \textcolor{myred}{red boxes}. In the rest detection results, \textcolor{myred}{red circles} denote the \textcolor{myred}{TP} detections; \textcolor{mygreen}{green circles} denote the \textcolor{mygreen}{FP} detections; and \textcolor{myorange}{orange circles} denote the \textcolor{myorange}{FN} detections.}
    \label{fig:nudtVisual}
\end{figure*}

\begin{figure*}[!t]
    \centering
    \includegraphics[width=16cm]{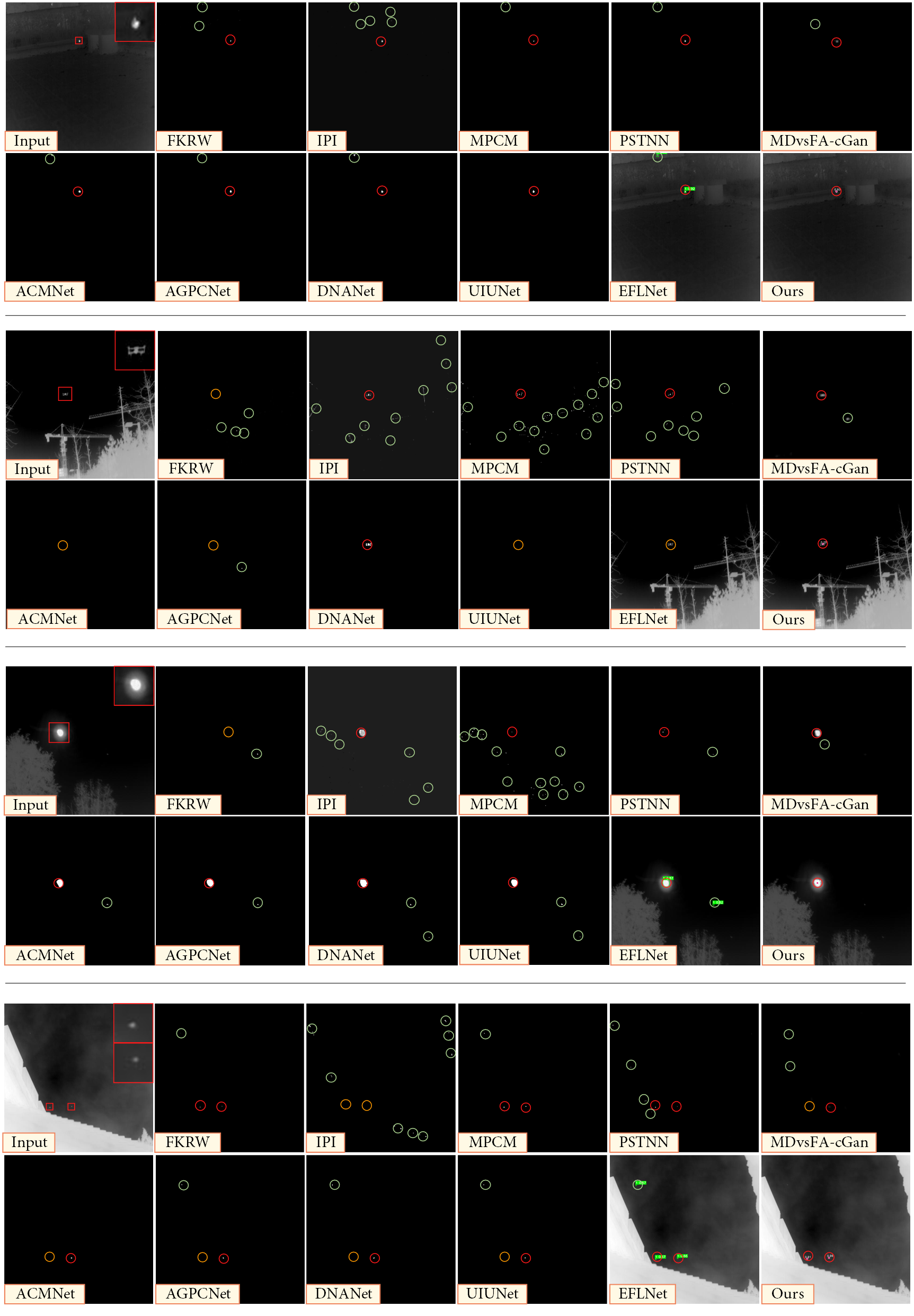}
    \caption{Visual results of different methods on the IRSTD-1k dataset. In the input images, real targets are indicated by the \textcolor{myred}{red boxes}. In the rest detection results, \textcolor{myred}{red circles} denote the \textcolor{myred}{TP} detections; \textcolor{mygreen}{green circles} denote the \textcolor{mygreen}{FP} detections; and \textcolor{myorange}{orange circles} denote the \textcolor{myorange}{FN} detections.}
    \label{fig:1kVisual}
\end{figure*}

To demonstrate the excellent performance of SSHD-Net, we perform comparison experiments with other methods on the NUDT-SIRST \cite{li2022DNA} and IRSTD-1k\cite{zhang2022isnet_msaff16} datasets, and the visualization results are shown in \Cref{fig:nudtVisual,fig:1kVisual}, respectively. In the input images, \textcolor{myred}{real targets} are marked with \textcolor{myred}{red boxes}. In the detection results of each method, the \textcolor{myred}{red circles} represent \textcolor{myred}{true positives (TP)}, \textcolor{mygreen}{green circles} represent \textcolor{mygreen}{false positives (FP)}, and \textcolor{myorange}{orange circles} represent \textcolor{myorange}{false negatives (FN)}. The modules with superior performance have more \textcolor{myred}{red circles} and fewer \textcolor{mygreen}{green} and \textcolor{myorange}{orange} circles. Experiments show that M-based methods\cite{qin2019FKRW_qiu120}\cite{gao2013IPI_efl8}\cite{wei2016MPCM_qiu110}\cite{zhang2019PSTNN_efl24}, although capable of detecting most targets, are prone to generating a large number of false alarms under complex backgrounds and noise interference owing to over-reliance on a priori assumptions. In contrast, DL-based methods\cite{wang2019MDvsFA_efl10}\cite{dai2021acm_efl11}\cite{zhang2023agpc}\cite{li2022DNA}\cite{wu2022uiu} \cite{yang2024eflnet} significantly reduce false alarms. However, in situations with a bright background around the target or dimly lit targets, these methods are susceptible to background interference, leading to false alarms and missed detections. This is due to the blurring of the target edges by the thermal-effect of infrared radiation, making it difficult to clearly distinguish the target from the background. Consequently, this inaccuracy in labelling leads to the incorrect allocation of positive and negative samples, which affects the detection accuracy. In addition, the small size of IR small-target makes it difficult for existing networks to strike a balance between the feature map size and network depth, which is insufficient for fully learning the target information, thus affecting the detection effect in real complex scenes. In contrast, SSHD-Net effectively adapts to the challenges of various cluttered backgrounds, target shapes and sizes by maintaining high-resolution features and employing techniques such as dynamic feature enhancement. This enables the network to demonstrate superior performance in complex scenarios.

\begin{table}[]
\caption{Ablation Study On SSHD-Net's Choice Of Feature Extraction Network Width\label{tab:Network Width}}
\centering
\renewcommand{\arraystretch}{1.5} 
\setlength{\tabcolsep}{5mm}
\begin{tabular}{c|ccc}
\thickhline
\textbf{Models}      & {\color[HTML]{393939} \textbf{Pre}}   & {\color[HTML]{393939} \textbf{Rec}}   & {\color[HTML]{393939} \textbf{F1}}    \\ \hline
SSHD-Net-W24          & {\color[HTML]{393939} 94.16}          & {\color[HTML]{393939} 88.05}          & {\color[HTML]{393939} 91.01}          \\
\textbf{SSHD-Net-W32} & {\color[HTML]{393939} \textbf{96.42}} & {\color[HTML]{393939} \textbf{92.76}} & {\color[HTML]{393939} \textbf{94.55}} \\
SSHD-Net-W48          & {\color[HTML]{393939} 91.92}          & {\color[HTML]{393939} 81.57}          & {\color[HTML]{393939} 86.44}          \\ \thickhline
\end{tabular}
\end{table}

\begin{table}[]
\caption{Ablation Study On The Different Hyperparameter $\lambda$ Form Of Adaptive Target localization Detection Head \label{tab:ab ATLDH}}
\centering
\renewcommand{\arraystretch}{1.5} 
\setlength{\tabcolsep}{5mm}
\begin{tabular}{c|ccc}
\thickhline
\textbf{Hyperparameter $\lambda$} & {\color[HTML]{393939} \textbf{Pre}}   & {\color[HTML]{393939} \textbf{Rec}}   & {\color[HTML]{393939} \textbf{F1}}    \\ \hline
{\color[HTML]{393939} 0.05}                     & {\color[HTML]{393939} 97.27}          & {\color[HTML]{393939} 85.86}          & {\color[HTML]{393939} 91.21}          \\
{\color[HTML]{393939} 0.10}                     & {\color[HTML]{393939} 97.42}          & {\color[HTML]{393939} 91.03}          & {\color[HTML]{393939} 94.12}          \\
{\color[HTML]{393939} 0.15}                     & {\color[HTML]{393939} \textbf{97.43}}          & {\color[HTML]{393939} 91.38}          & {\color[HTML]{393939} 94.31}          \\
{\color[HTML]{393939} 0.20}                     & {\color[HTML]{393939} 96.75}          & {\color[HTML]{393939} 92.41}          & {\color[HTML]{393939} 94.53}          \\
{\color[HTML]{393939} 0.25}                     & {\color[HTML]{393939} 96.42} & {\color[HTML]{393939} \textbf{92.76}} & {\color[HTML]{393939} \textbf{94.55}} \\
{\color[HTML]{393939} 0.30}                     & {\color[HTML]{393939} 96.07}          & {\color[HTML]{393939} 92.76}          & {\color[HTML]{393939} 94.39}          \\
{\color[HTML]{393939} 0.35}                     & {\color[HTML]{393939} 96.07}          & {\color[HTML]{393939} 92.76}          & {\color[HTML]{393939} 94.39}          \\
{\color[HTML]{393939} 0.40}                     & {\color[HTML]{393939} 95.73}          & {\color[HTML]{393939} 92.76}          & {\color[HTML]{393939} 94.22}          \\
{\color[HTML]{393939} 0.45}                     & {\color[HTML]{393939} 95.39}          & {\color[HTML]{393939} 92.76}          & {\color[HTML]{393939} 94.06}          \\
{\color[HTML]{393939} 0.50}                     & {\color[HTML]{393939} 95.05}          & {\color[HTML]{393939} 92.76}          & {\color[HTML]{393939} 93.89}          \\
{\color[HTML]{393939} 0.55}                     & {\color[HTML]{393939} 94.72}          & {\color[HTML]{393939} 92.76}          & {\color[HTML]{393939} 93.73}          \\ 
\thickhline
\end{tabular}
\end{table}

\begin{figure*} 
    \centering
    \includegraphics[width=\textwidth]{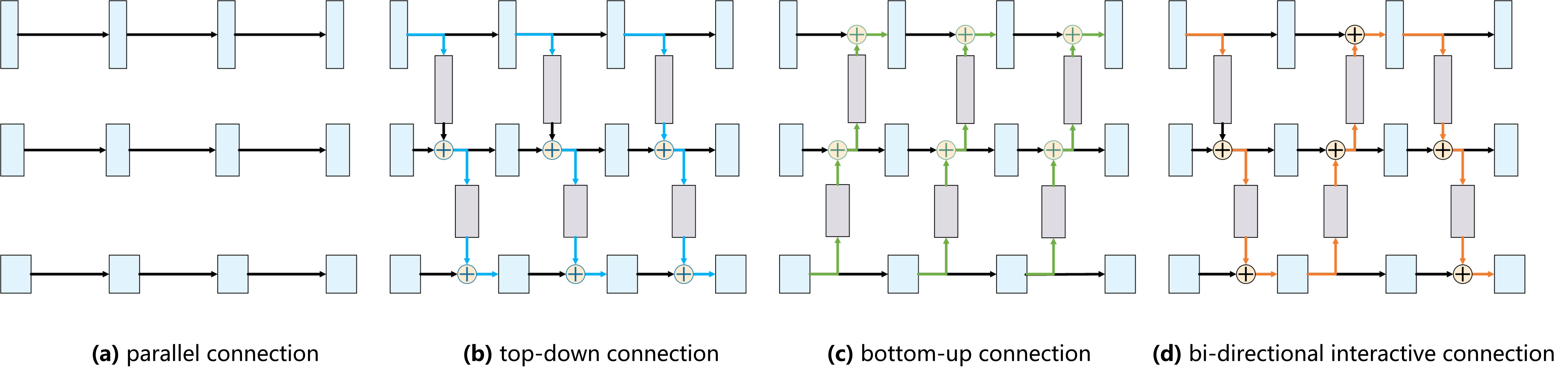}
    \caption{Different variants in HCEM ablation experiments. (a) SSHD-Net with parallel connections. (b) SSHD-Net with top-down connections. (c) SSHD-Net with bottom-up connections. (d) SSHD-Net with HCEM bi-directional interaction connections.}
    \label{fig:Impact Of HCEM}
\end{figure*}

\begin{table}[]
\caption{Ablation Study For Impact Of High-Resolution Cross-Feature Extraction Module\label{tab:Impact Of HCEM}}
\centering
\renewcommand{\arraystretch}{1.5} 
\setlength{\tabcolsep}{5mm}
\begin{tabular}{c|ccc}
\thickhline
\textbf{Models}   & {\color[HTML]{393939} \textbf{Pre}}   & {\color[HTML]{393939} \textbf{Rec}}   & {\color[HTML]{393939} \textbf{F1}}    \\ \hline
SSHD-Net-bottom-up & {\color[HTML]{393939} 94.87}          & {\color[HTML]{393939} 88.40}          & {\color[HTML]{393939} 91.52}          \\
SSHD-Net w/o HCEM  & {\color[HTML]{393939} 95.60}          & {\color[HTML]{393939} 89.08}          & {\color[HTML]{393939} 92.23}          \\
SSHD-Net-top-down  & {\color[HTML]{393939} 95.31}          & {\color[HTML]{393939} 90.10}          & {\color[HTML]{393939} 92.63}          \\
\textbf{SSHD-Net w/ HCEM}  & {\color[HTML]{393939} \textbf{96.42}} & {\color[HTML]{393939} \textbf{92.76}} & {\color[HTML]{393939} \textbf{94.55}} \\ \thickhline
\end{tabular}
\end{table}

\begin{table}[]
\caption{Ablation Study For Impact Of Dynamic Coordinate Fusion Module\label{tab:Impact Of DCFM}}
\centering
\renewcommand{\arraystretch}{1.5} 
\setlength{\tabcolsep}{2.5mm}
\begin{tabular}{c|ccc}
\thickhline
\textbf{Models}                 & {\color[HTML]{393939} \textbf{Pre}}   & {\color[HTML]{393939} \textbf{Rec}}   & {\color[HTML]{393939} \textbf{F1}}    \\ \hline
SSHD-Net w/o DCFM(part b-I\&part b-II) & {\color[HTML]{393939} 93.51}          & {\color[HTML]{393939} 83.62}          & {\color[HTML]{393939} 88.29}          \\
SSHD-Net w/o DCFM(part b-II)        & {\color[HTML]{393939} 92.62}          & {\color[HTML]{393939} 85.67}          & {\color[HTML]{393939} 89.01}          \\
SSHD-Net w/o DCFM(part b-I)         & {\color[HTML]{393939} 94.87}          & {\color[HTML]{393939} 88.40}          & {\color[HTML]{393939} 91.52}          \\
SSHD-Net w/o CA                  & {\color[HTML]{393939} 94.16}          & {\color[HTML]{393939} 88.05}          & {\color[HTML]{393939} 91.01}          \\
SSHD-Net w/o ODBlock             & {\color[HTML]{393939} 93.99}          & {\color[HTML]{393939} 90.78}          & {\color[HTML]{393939} 92.36}          \\
\textbf{SSHD-Net w/ DCFM(part b-I\textup{\&}part b-II)}                & {\color[HTML]{393939} \textbf{96.42}} & {\color[HTML]{393939} \textbf{92.76}} & {\color[HTML]{393939} \textbf{94.55}} \\ \thickhline
\end{tabular}
\end{table}

\begin{figure} 
    \centering
    \includegraphics[width=\linewidth]{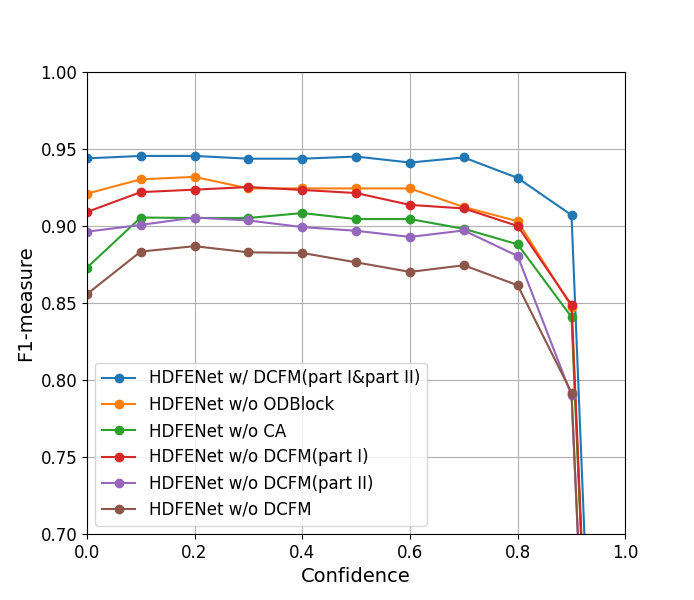}
    \caption{F1-Confidence line graph in the DCFM ablation experiment. The graph demonstrates the importance of DCFM in the network.}
    \label{fig:F1-confidence}
\end{figure}

\subsection{Ablation Study}

To validate the effectiveness of each module in SSHD-Net, we conducted a series of ablation experiments. We selected the IRSTD-1k dataset as the basis for the experiments, which contains numerous realistic and complex scenes that pose significant challenges to the IRSTD task.
\begin{enumerate}[wide] 

\item{\textit{\textbf{Choice of Feature Extraction Network Width}}:}
In the IRSTD task, increasing the network width can capture richer target features but may also introduce more background noise, which makes sparse IR small-target features difficult to identify. Therefore, comparative experiments are conducted by setting different network widths to balance the network performance and efficiency. The experimental comparison results of SSHD-Net with different network widths are listed in \Cref{tab:Network Width}. Where SSHD-Net-W24, SSHD-Net-W32, and SSHD-Net-W48 denote that the number of channels of ${\mathbb{F}}_{j}^{i}$, ${\mathbb{L}}_{j}^{i}$, and ${\mathbb{H}}_{j}^{i}$ are $j\times24$,  $j\times32$, and $j\times48$, respectively.

The results demonstrate that SSHD-Net-W24 decreases by 2.26$\%$, 4.71$\%$, and 3.54$\%$ in Pre, Rec, and F1, respectively, compared to SSHD-Net-W32. This suggests that an insufficient network width may lead to poor feature representation and difficulty in effectively distinguishing between the target and background. In contrast, although SSHD-Net-W48 increases the network width, it generate a large amount of background noise that overwhelms the sparse small target features, causing the Pre, Rec and F1 to decrease by 4.50$\%$, 11.19$\%$ and 8.11$\%$, respectively. This result suggests that excessively wide networks are detrimental to improving detection performance. In conclusion, we select SSHD-Net-W32 for subsequent experiments.

\item{\textit{\textbf{Choice of Hyperparamter}} $\lambda$:} In order to bring out the best performance of our ATLDH, we set multiple sets of hyperparameters $\lambda$ for comparison experiments, and the specific results are shown in \Cref{tab:ab ATLDH}. It is found that when the value of $\lambda$ is low, ATLDH is trying to control the target confidence excessively strictly, which slightly improves the Pre but significantly reduces the Rec, resulting in a relatively low F1. Conversely, when the value of $\lambda$ is large, ATLDH is weakened in its ability to suppress the background, leading to a decrease in precison.

The experimental results show that when the hyperparameter $\lambda$ is set to 0.25, the ATLDH can achieve optimal performance and obtain optimal detection results. This hyperparameter setting effectively balances strict control of the target confidence with the ability to suppress the background and achieve an optimal combination of accuracy and regression. Therefore, we select this parameter setting for subsequent experiments.

\item{\textit{\textbf{Impact of High-Resolution Cross-Feature Extraction Module (HCEM)}}:} The high-resolution cross-feature extraction module maintains parallel propagation of high- and low-resolution feature layers during the feature propagation process, effectively avoiding the loss of IR small-target information in the deep network. In addition, through a stepped feature cascade channel (SFCC), HCEM achieves adaptive contextual feature extraction with progressive bi-directional interaction, which optimizes the information flow between feature layers. To demonstrate the effectiveness of our HCEM, we design three network variants and conduct multiple sets of experiments to obtain the mean value to ensure fairness and accuracy of the experimental results. \Cref{tab:Impact Of HCEM} shows the results of the SSHD-Net with HCEM compared with the other variants in the experiments. The results show that SSHD-Net with parallel connections decreases by 0.44$\%$, 2.67$\%$ and 1.61$\%$ for Pre, Rec and F1, respectively. This indicates that the HCEM efficiently extracts multiple semantic information of the target by effectively fusing multi-scale interaction features.

\setlist[itemize]{leftmargin=2em}
\begin{itemize}
  \item{\textbf{SSHD-Net w/o HCEM}:} as illustrated in \Cref{fig:Impact Of HCEM} (a), we remove the SFCC and use parallel propagation to extract the features directly at different scales.
  \item{\textbf{SSHD-Net-bottom-up}:} In this variant, we design all the cascade channels in a bottom-up manner, allowing low-resolution features to be fed back into the high-resolution feature layer, as illustrated in \Cref{fig:Impact Of HCEM} (b).
  \item{\textbf{SSHD-Net-top-down}:} as illustrated in \Cref{fig:Impact Of HCEM} (c), this variant uses a top-down feature fusion cascade channel, where information from the high-resolution feature layer cascades guides the extraction of the low-resolution feature layer.
\end{itemize}

As shown in \Cref{tab:Impact Of HCEM}, the Pre, Rec and F1 of SSHD-Net-bottom-up on the dataset decreased by 1.55$\%$, 4.36$\%$ and 3.03$\%$, respectively. This result suggests that although bottom-up connectivity increases the flow of information from the low-resolution feature layer to the high-resolution feature layer, this method ignores the importance of the target details in the high-resolution layer for overall feature fusion. In contrast, SSHD-Net w/ HCEM utilizes the rich target information contained in the high-resolution layer to guide the low-resolution feature layer, which makes more effective use of the IR small-target feature information. Thus, the results indicate that the guiding role of high-resolution features is crucial for ensuring IRSTD precision.

Compared with the SSHD-Net containing the high-resolution cross-feature enhancement module (HCEM), Pre, Rec and F1 of the variant SSHD-Net-top-down decreased by 1.11$\%$, 2.66$\%$ and 1.92$\%$, respectively. The results indicate that although top-down connectivity effectively maintains the transfer of small target information in the high-resolution layer, it fails to fully utilize the global feature information in the low-resolution layer. This prevents the network from understanding the feature classes better. Consequently, it is critical to effectively balance local details with global contextual information in the network.

\item{\textit{\textbf{Impact of Dynamic Coordinate Fusion Module (DCFM)}} :} 
We propose a dynamic coordinate fusion module that significantly enhances the feature extraction capability of the network by dynamically integrating multiple types of contextual information and focusing on the target localization information. To validate the effectiveness and innovativeness of the DCFM, we design five variants to compare multiple sets of experiments and used the mean value as the final evaluation result.

\setlist[itemize]{leftmargin=2em}
\begin{itemize}
  \item{\textbf{SSHD-Net w/o DCFM}:} In this variant, we remove the DCFM module from the entire network and use the conventional feature map addition operation for the fusion of features of different resolutions.
  \item{\textbf{SSHD-Net w/o DCFM(part b-I) $\textup{\&}$ w/o DCFM(part b-II)}:} In these two variants, we remove the DCFM modules in parts b-I and b-II, as illustrated in \Cref{fig:Overall_Architecture}. We then replace the removed feature fusion part with a feature map addition operation.
  \item{\textbf{SSHD-Net w/o ODBlock}:} In this variant, we remove the ODBlock module from the DCFM to evaluate the effect of ODBlock.
  \item{\textbf{SSHD-Net w/o CA}:} We remove the CA  module from the DCFM in this variant to investigate the benefits of the CA  module.
\end{itemize}

\Cref{fig:F1-confidence} shows the F1 curves for the SSHD-Net and its five variants with confidence thresholds. From the figure, it can be seen that the six models perform relatively stable when the confidence threshold is in the 0.1-0.7 interval; in the 0.8-1.0 interval, the performance drops sharply. This suggests that an excessively high confidence threshold will lose some of the correct predictions, resulting in a lower F1. To ensure the fairness of the comparison experiment, we set the confidence threshold to 0.2 for the following experiments, and the results of the evaluation index comparison are shown in \Cref{tab:Impact Of DCFM}.

When the DCFM is completely removed, the network performs the worst. The Pre, Rec, and F1 decrease by 2.91$\%$, 9.14$\%$, and 6.26$\%$, respectively, compared with SSHD-Net w/ DCFM. This proves the critical role of the DCFM in fusing multiple semantic information with learning the target location information.

When the network removes the DCFM (part-I) and DCFM (part-II) modules, the Pre decreases by 1.55$\%$, and 3.80$\%$, Rec decreases by 4.36$\%$, 7.09$\%$, and F1 decreases by 3.03$\%$, and 5.54$\%$, respectively. This further demonstrates that DCFM can play an active role at any depth of the network, efficiently achieving multi-scale feature fusion.

In the variant with the ODBlock removed, the Pre, Rec, and F1 decreased by 2.43$\%$, 1.98$\%$, and 2.19$\%$, respectively. This is because ODBlock produces excellent results by enabling the network to focus on critical information and suppressing complex background interference.

When the CA module is removed, the number of variants in Pre, Rec and F1 decrease by 2.26$\%$, 4.71$\%$ and 3.54$\%$, respectively. This result shows that the CA module can effectively improve the attention of the network to the local target location information and prevent the target information from being overwhelmed by background interference after feature fusion.

\end{enumerate}

\section{Conclusion}
\label{sec:Conclusion}
In this study, we design a single-point supervised high-resolution dynamic network (SSHD-Net) to overcome several challenges in IRSTD tasks, using point-to-point target regression. Using the high-resolution cross-feature extraction module (HCEM), we maintain the IR small-target information in the deep network and achieve a balance between network depth and feature resolution. The effective integration of multiple special features is achieved by implementing a dynamic coordinate fusion module (DCFM). In addition, the creative design of the adaptive target localization detection head (ATLDH), which employs the SPTG as the training supervisory signal, enables the model to focus on the target core area more accurately. Experimental results on two challenging public datasets show that SSHD-Net is better than the SOTA methods in terms of IRSTD performance with only a single point of supervision.

\bibliographystyle{IEEEtran}
\bibliography{Mybib}

\begin{thebibliography}{10}
\providecommand{\url}[1]{#1}
\csname url@samestyle\endcsname
\providecommand{\newblock}{\relax}
\providecommand{\bibinfo}[2]{#2}
\providecommand{\BIBentrySTDinterwordspacing}{\spaceskip=0pt\relax}
\providecommand{\BIBentryALTinterwordstretchfactor}{4}
\providecommand{\BIBentryALTinterwordspacing}{\spaceskip=\fontdimen2\font plus
\BIBentryALTinterwordstretchfactor\fontdimen3\font minus \fontdimen4\font\relax}
\providecommand{\BIBforeignlanguage}[2]{{%
\expandafter\ifx\csname l@#1\endcsname\relax
\typeout{** WARNING: IEEEtran.bst: No hyphenation pattern has been}%
\typeout{** loaded for the language `#1'. Using the pattern for}%
\typeout{** the default language instead.}%
\else
\language=\csname l@#1\endcsname
\fi
#2}}
\providecommand{\BIBdecl}{\relax}
\BIBdecl

\bibitem{Missile_Defense_RLPGB_2}
R.~C. Hall, \emph{Missile Defense Alarm: The Genesis of Space-based Infrared Early Warning}.\hskip 1em plus 0.5em minus 0.4em\relax NRO History Office, 1999.

\bibitem{sun2020infrared_EFL_3}
Y.~Sun, J.~Yang, and W.~An, ``Infrared dim and small target detection via multiple subspace learning and spatial-temporal patch-tensor model,'' \emph{IEEE Transactions on Geoscience and Remote Sensing}, vol.~59, no.~5, pp. 3737--3752, 2020.

\bibitem{ying2022mocopnet_DNA_2}
X.~Ying, Y.~Wang, L.~Wang, W.~Sheng, L.~Liu, Z.~Lin, and S.~Zhou, ``Mocopnet: Exploring local motion and contrast priors for infrared small target super-resolution,'' \emph{arXiv e-prints}, pp. arXiv--2201, 2022.

\bibitem{teutsch2010classification_DNA_1}
M.~Teutsch and W.~Kr{\"u}ger, ``Classification of small boats in infrared images for maritime surveillance,'' in \emph{2010 International WaterSide Security Conference}.\hskip 1em plus 0.5em minus 0.4em\relax IEEE, 2010, pp. 1--7.

\bibitem{wang2023rlpgb}
Z.~Wang, T.~Zang, Z.~Fu, H.~Yang, and W.~Du, ``Rlpgb-net: Reinforcement learning of feature fusion and global context boundary attention for infrared dim small target detection,'' \emph{IEEE Transactions on Geoscience and Remote Sensing}, 2023.

\bibitem{zhang2003algorithms_RLPGB_10}
W.~Zhang, M.~Cong, and L.~Wang, ``Algorithms for optical weak small targets detection and tracking,'' in \emph{International Conference on Neural Networks and Signal Processing, 2003. Proceedings of the 2003}, vol.~1.\hskip 1em plus 0.5em minus 0.4em\relax IEEE, 2003, pp. 643--647.

\bibitem{li2022DNA}
B.~Li, C.~Xiao, L.~Wang, Y.~Wang, Z.~Lin, M.~Li, W.~An, and Y.~Guo, ``Dense nested attention network for infrared small target detection,'' \emph{IEEE Transactions on Image Processing}, vol.~32, pp. 1745--1758, 2022.

\bibitem{zhang2014thermaleffectofinfrared}
C.-s. Zhang, Z.~Shi, B.~Feng, and B.-s. Xu, ``Infrared lens thermal effect: equivalent focal shift and calculating model,'' in \emph{International Symposium on Optoelectronic Technology and Application 2014: Infrared Technology and Applications}, vol. 9300.\hskip 1em plus 0.5em minus 0.4em\relax SPIE, 2014, pp. 451--458.

\bibitem{yang2024eflnet}
B.~Yang, X.~Zhang, J.~Zhang, J.~Luo, M.~Zhou, and Y.~Pi, ``Eflnet: Enhancing feature learning network for infrared small target detection,'' \emph{IEEE Transactions on Geoscience and Remote Sensing}, vol.~62, pp. 1--11, 2024.

\bibitem{deshpande1999max_efl4}
S.~D. Deshpande, M.~H. Er, R.~Venkateswarlu, and P.~Chan, ``Max-mean and max-median filters for detection of small targets,'' in \emph{Signal and Data Processing of Small Targets 1999}, vol. 3809.\hskip 1em plus 0.5em minus 0.4em\relax SPIE, 1999, pp. 74--83.

\bibitem{wang2020wavelet_qiu86}
H.~Wang and Y.~Xin, ``Wavelet-based contourlet transform and kurtosis map for infrared small target detection in complex background,'' \emph{Sensors}, vol.~20, no.~3, p. 755, 2020.

\bibitem{hussain1995infrared_qiu3}
T.~M. Hussain, A.~M. Baig, T.~N. Saadawi, and S.~A. Ahmed, ``Infrared pyroelectric sensor for detection of vehicular traffic using digital signal processing techniques,'' \emph{IEEE transactions on vehicular technology}, vol.~44, no.~3, pp. 683--689, 1995.

\bibitem{wei2016MPCM_qiu110}
Y.~Wei, X.~You, and H.~Li, ``Multiscale patch-based contrast measure for small infrared target detection,'' \emph{Pattern Recognition}, vol.~58, pp. 216--226, 2016.

\bibitem{qin2019FKRW_qiu120}
Y.~Qin, L.~Bruzzone, C.~Gao, and B.~Li, ``Infrared small target detection based on facet kernel and random walker,'' \emph{IEEE Transactions on Geoscience and Remote Sensing}, vol.~57, no.~9, pp. 7104--7118, 2019.

\bibitem{qiu2020adaptive_qiu_Self1}
Z.~Qiu, Y.~Ma, F.~Fan, J.~Huang, and M.~Wu, ``Adaptive scale patch-based contrast measure for dim and small infrared target detection,'' \emph{IEEE Geoscience and Remote Sensing Letters}, vol.~19, pp. 1--5, 2022.

\bibitem{gao2013IPI_efl8}
C.~Gao, D.~Meng, Y.~Yang, Y.~Wang, X.~Zhou, and A.~G. Hauptmann, ``Infrared patch-image model for small target detection in a single image,'' \emph{IEEE transactions on image processing}, vol.~22, no.~12, pp. 4996--5009, 2013.

\bibitem{dai2017reweighted_efl23}
Y.~Dai and Y.~Wu, ``Reweighted infrared patch-tensor model with both nonlocal and local priors for single-frame small target detection,'' \emph{IEEE journal of selected topics in applied earth observations and remote sensing}, vol.~10, no.~8, pp. 3752--3767, 2017.

\bibitem{zhang2019PSTNN_efl24}
L.~Zhang and Z.~Peng, ``Infrared small target detection based on partial sum of the tensor nuclear norm,'' \emph{Remote Sensing}, vol.~11, no.~4, p. 382, 2019.

\bibitem{dai2021alcnet}
Y.~Dai, Y.~Wu, F.~Zhou, and K.~Barnard, ``Attentional local contrast networks for infrared small target detection,'' \emph{IEEE transactions on geoscience and remote sensing}, vol.~59, no.~11, pp. 9813--9824, 2021.

\bibitem{liu2023infrared_kaiti39}
F.~Liu, C.~Gao, F.~Chen, D.~Meng, W.~Zuo, and X.~Gao, ``Infrared small and dim target detection with transformer under complex backgrounds,'' \emph{IEEE Transactions on Image Processing}, vol.~32, pp. 5921--5932, 2023.

\bibitem{chen2022irstformer_efl29}
G.~Chen, W.~Wang, and S.~Tan, ``Irstformer: A hierarchical vision transformer for infrared small target detection,'' \emph{Remote Sensing}, vol.~14, no.~14, p. 3258, 2022.

\bibitem{wang2019MDvsFA_efl10}
H.~Wang, L.~Zhou, and L.~Wang, ``Miss detection vs. false alarm: Adversarial learning for small object segmentation in infrared images,'' in \emph{Proceedings of the IEEE/CVF International Conference on Computer Vision}, 2019, pp. 8509--8518.

\bibitem{dai2021acm_efl11}
Y.~Dai, Y.~Wu, F.~Zhou, and K.~Barnard, ``Asymmetric contextual modulation for infrared small target detection,'' in \emph{Proceedings of the IEEE/CVF Winter Conference on Applications of Computer Vision}, 2021, pp. 950--959.

\bibitem{dai2023oscar_self}
Y.~Dai, X.~Li, F.~Zhou, Y.~Qian, Y.~Chen, and J.~Yang, ``One-stage cascade refinement networks for infrared small target detection,'' \emph{IEEE Transactions on Geoscience and Remote Sensing}, vol.~61, pp. 1--17, 2023.

\bibitem{ying2023LESPS}
X.~Ying, L.~Liu, Y.~Wang, R.~Li, N.~Chen, Z.~Lin, W.~Sheng, and S.~Zhou, ``Mapping degeneration meets label evolution: Learning infrared small target detection with single point supervision,'' in \emph{Proceedings of the IEEE/CVF Conference on Computer Vision and Pattern Recognition}, 2023, pp. 15\,528--15\,538.

\bibitem{zhang2022isnet_msaff16}
M.~Zhang, R.~Zhang, Y.~Yang, H.~Bai, J.~Zhang, and J.~Guo, ``Isnet: Shape matters for infrared small target detection,'' in \emph{Proceedings of the IEEE/CVF Conference on Computer Vision and Pattern Recognition}, 2022, pp. 877--886.

\bibitem{bai2010tophat_rlpgb59}
X.~Bai and F.~Zhou, ``Analysis of new top-hat transformation and the application for infrared dim small target detection,'' \emph{Pattern Recognition}, vol.~43, no.~6, pp. 2145--2156, 2010.

\bibitem{qiu2022PLLCM}
Z.~Qiu, Y.~Ma, F.~Fan, J.~Huang, M.~Wu, and X.~Mei, ``A pixel-level local contrast measure for infrared small target detection,'' \emph{Defence Technology}, vol.~18, no.~9, pp. 1589--1601, 2022.

\bibitem{chen2024rumfr}
L.~Chen, T.~Wu, S.~Zheng, Z.~Qiu, and F.~Huang, ``Robust unsupervised multi-feature representation for infrared small target detection,'' \emph{IEEE Journal of Selected Topics in Applied Earth Observations and Remote Sensing}, 2024.

\bibitem{mirza2014cGan}
M.~Mirza and S.~Osindero, ``Conditional generative adversarial nets,'' \emph{arXiv preprint arXiv:1411.1784}, 2014.

\bibitem{du2021spatial_kaiti31}
J.~Du, H.~Lu, L.~Zhang, M.~Hu, S.~Chen, Y.~Deng, X.~Shen, and Y.~Zhang, ``A spatial-temporal feature-based detection framework for infrared dim small target,'' \emph{IEEE Transactions on Geoscience and Remote Sensing}, vol.~60, pp. 1--12, 2021.

\bibitem{ju2021istdet_kaiti32}
M.~Ju, J.~Luo, G.~Liu, and H.~Luo, ``Istdet: An efficient end-to-end neural network for infrared small target detection,'' \emph{Infrared Physics \& Technology}, vol. 114, p. 103659, 2021.

\bibitem{zhou2022yolo_kaiti33}
X.~Zhou, L.~Jiang, C.~Hu, S.~Lei, T.~Zhang, and X.~Mou, ``Yolo-sase: An improved yolo algorithm for the small targets detection in complex backgrounds,'' \emph{Sensors}, vol.~22, no.~12, p. 4600, 2022.

\bibitem{he2016resnet}
K.~He, X.~Zhang, S.~Ren, and J.~Sun, ``Deep residual learning for image recognition,'' in \emph{Proceedings of the IEEE conference on computer vision and pattern recognition}, 2016, pp. 770--778.

\bibitem{howard2017mobilenets}
A.~G. Howard, M.~Zhu, B.~Chen, D.~Kalenichenko, W.~Wang, T.~Weyand, M.~Andreetto, and H.~Adam, ``Mobilenets: Efficient convolutional neural networks for mobile vision applications,'' \emph{arXiv preprint arXiv:1704.04861}, 2017.

\bibitem{sun2019hrent}
K.~Sun, B.~Xiao, D.~Liu, and J.~Wang, ``Deep high-resolution representation learning for human pose estimation,'' in \emph{Proceedings of the IEEE/CVF conference on computer vision and pattern recognition}, 2019, pp. 5693--5703.

\bibitem{li2022odconv}
C.~Li, A.~Zhou, and A.~Yao, ``Omni-dimensional dynamic convolution,'' \emph{arXiv preprint arXiv:2209.07947}, 2022.

\bibitem{hou2021coordinateAtt}
Q.~Hou, D.~Zhou, and J.~Feng, ``Coordinate attention for efficient mobile network design,'' in \emph{Proceedings of the IEEE/CVF conference on computer vision and pattern recognition}, 2021, pp. 13\,713--13\,722.

\bibitem{zhang2023agpc}
T.~Zhang, L.~Li, S.~Cao, T.~Pu, and Z.~Peng, ``Attention-guided pyramid context networks for detecting infrared small target under complex background,'' \emph{IEEE Transactions on Aerospace and Electronic Systems}, 2023.

\bibitem{wu2022uiu}
X.~Wu, D.~Hong, and J.~Chanussot, ``Uiu-net: U-net in u-net for infrared small object detection,'' \emph{IEEE Transactions on Image Processing}, vol.~32, pp. 364--376, 2022.

\end{thebibliography}




\end{document}